\documentclass[make,article,submit,pdftex,moreauthors]{Definitions/mdpi}

\usepackage[labelformat=simple]{subcaption}

\DeclareCaptionLabelFormat{subcaptionlabel}{\normalfont(\textbf{#2}\normalfont)}
\usepackage{multicol}
\usepackage{changes}
\usepackage{balance}
\usepackage[nolist]{acronym}
\usepackage{pgfplots}
\usepackage{multirow}
\usepackage{longtable}
\usepackage{pifont}
\newcommand{\cmark}{\ding{51}}%
\newcommand{\xmark}{\text{\ding{55}}}

\firstpage{1} 
\pubvolume{1}
\issuenum{1}
\articlenumber{0}
\pubyear{2025}
\copyrightyear{2025}
\externaleditor{Firstname Lastname}
\datereceived{13 April 2025} 
\daterevised{} 
\dateaccepted{} 
\datepublished{ } 
\hreflink{https://doi.org/} 

\Title{Machine Learning and Transformers for Thyroid Carcinoma Diagnosis}

\TitleCitation{Machine Learning and Transformers for Thyroid Carcinoma Diagnosis}

\Author{Yassine Habchi  \orcidA{}, Hamza  Kheddar \orcidB{}, Yassine Himeur\orcidC{}, Mohamed Chahine Ghanem\orcidD{} }

\AuthorNames{Yassine Habchi, Hamza Kheddar, Yassine Himeur, Mohamed Chahine Ghanem}

\AuthorCitation{Habchi, et al.}

\address{%
$^{1}$ \quad Institute of Technology, University Center Salhi Ahmed, Naama, Algeria.\\
$^{2}$ \quad LSEA Laboratory, Electrical Engineering Department, University of Medea, 26000, Algeria.\\
$^{3}$ \quad College of Engineering and Information Technology, University of Dubai, Dubai, UAE. \\
$^{4}$ \quad Cybersecurity Institute, Department of Computer Science, University of Liverpool, Liverpool, UK.
}

\corres{\hangafter=1 \hangindent=1.05em \hspace{-0.82em} Correspondence: mohamed.chahine.ghanem@liverpool.ac.uk 
}

\abstract{The growing interest in developing smart diagnostic systems to help medical experts process extensive data for treating incurable diseases has been notable. In particular, the challenge of identifying thyroid cancer (TC) has seen progress with the use of machine learning (ML) and big data analysis, incorporating Transformers to evaluate TC prognosis and determine the risk of malignancy in individuals. This review article presents a summary of various studies on AI-based approaches, especially those employing Transformers, for diagnosing TC. It introduces a new categorization system for these methods based on artificial intelligence (AI) algorithms, the goals of the framework, and the computing environments used. Additionally, it scrutinizes and contrasts the available TC datasets by their features. The paper highlights the importance of AI instruments in aiding the diagnosis and treatment of TC through supervised, unsupervised, or mixed approaches, with a special focus on the ongoing importance of Transformers and large language models (LLMs) in medical diagnostics and disease management. It further discusses the progress made and the continuing obstacles in this area. Lastly, it explores future directions and focuses within this research field.}

\keyword{Thyroid carcinoma detection; Artificial intelligence; Deep learning; Biomedical images; Transformers; Large language model. } 

\begin{document}

\section{Introduction}  \label{intro}


The integration of \ac{AI} into the healthcare sector represents a pivotal advancement, fundamentally altering the landscape of medical diagnostics, therapy, and patient management. The superior capabilities of \ac{AI}, incorporating the identification of patterns, forecasting analytics, and the process of making decisions, have led to the creation of systems that can interpret intricate medical data with greater accuracy and scale than ever before \cite{himeur2023face,chouchane2023improving}. Such advancements facilitate the early identification of diseases, enhance the accuracy of diagnoses, and support the customization of treatment plans for individuals. In addition, predictive models powered by \ac{AI} are capable of foreseeing disease spread, boosting the efficiency of healthcare operations, and significantly improving outcomes for patients \cite{himeur2022deep}. \ac{AI} also has the potential to make healthcare more equitable by reducing disparities in service quality between rural and urban areas, thus improving access to premium healthcare services. As a result, the role of \ac{AI} is significant in healthcare and is anticipated to grow as ongoing technological innovations lead to the creation of even more advanced applications, promising widespread benefits for patient health across the globe \cite{sohail2023decoding,himeur2023ai}. 

Nonetheless, the trust serves as a critical intermediary, impacting the extent to which factors related to \ac{AI} affect user acceptance. Research has explored the roles of trust, risk, and security in determining the uptake of \ac{AI}-powered support \cite{calisto2022modeling}. Empirical investigations within these studies have underscored the essential influence of trust in forming the basis of user acceptance. Cancer is marked by the unchecked growth of cells across various parts of the body. These cells multiply erratically and can spread, damaging healthy tissue \cite{bechar2023harenessing}. Such uncontrolled cell growth is triggered by changes or mutations in the DNA of these cells \cite{habchi2023ai}. The DNA in cells comprises multiple genes, which provide the instructions necessary for a cell's function, growth, and division. When these instructions are erroneous, it can interrupt the normal functioning of cells and may result in the development of cancer \cite{salem2023cancer}. \Ac{TC}, in particular, is recognized as one of the most common types of endocrine malignancies around the globe \cite{deng2020global}.

Recent global epidemiological studies indicate a rise in abnormal \ac{TN}, linked to an upsurge in genetic cellular activity. This suggests an increase in normal cell functions, with anomalies classified into four primary types: \ac{FTC}, \ac{PTC}, \ac{MTC}, and \ac{ATC} \cite{castellana2020can,hitu2020microrna,giovanella2020eanm,ferrari2020novel}. Elements like exposure to radiation, Hashimoto's thyroiditis, psychological factors, and genetic components, alongside advances in technologies of detection, appear in these cancers. These factors can cause chronic health issues like diabetes and blood pressure instability. The cell cancer volume is key to evaluating the aggressiveness and prognosis of \ac{TC}'s, with cell nuclei detection offering alternative markers for evaluating cancer cell proliferation. \Ac{CAD} systems have gained prominence in \ac{TCD} analysis, improving diagnostic accuracy and reducing interpretation times \cite{khammari2023high}. Radiomics, particularly through \ac{US} imaging \cite{hamza2023hybrid}, has emerged as an efficient diagnostic method. The American College of Radiology's \ac{TIRADS} categorizes \acp{TN} from benign to malignant \cite{ tessler2018thyroid}. Despite available open-source tools for nodule analysis, accurately identifying them remains a challenge, reliant on radiologists' experience and the subjective nature of visual image analysis \cite{zhou2020differential}.

\begin{figure}[t!]
\centering
\includegraphics[width=0.8\columnwidth]{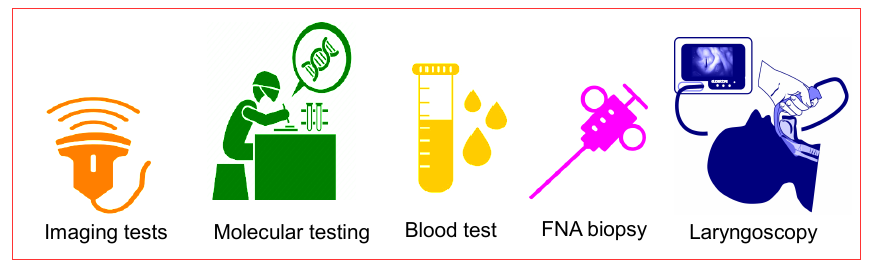}
\caption{Approaches for identifying \ac{TC}.}
\label{fig1}
\end{figure}

Additionally, \ac{US} imaging can be a lengthy and stress-inducing process, which may result in incorrect diagnoses. It is common to encounter classification errors among cases deemed normal, benign, malignant, or of uncertain nature \cite{nayak2020impact, kumar2020automated}. For a more precise diagnosis, a \ac{FNAB} is often conducted. Yet, this technique can be uncomfortable for patients, and inaccuracies by the practitioner can mistakenly label benign nodules as malignant, leading to unnecessary costs \cite{hahn2020comparison}. The main issue is the selection of nodule characteristics critical for accurately differentiating between benign and malignant cancer. Various research efforts have delved into the use of conventional \ac{US} imaging to characterize different types of cancers, such as retinal \cite{ullah2019ensemble}, breast \cite{ wang2023generalizable},  and thyroid \cite{wang2020comparison}. Despite these efforts, there remains a lack of accuracy in the methods available for effectively categorizing \acp{TN}, as depicted in  Figure \ref{fig1}.

The deployment of \ac{AI} technology is crucial in diminishing subjectivity and boosting the precision of pathological assessments, particularly for complex conditions like thyroid diseases \cite{zhang2019machine}. These advancements enhance the analysis of images obtained through \ac{US} and expedite analysis times. \Ac{ML} and \ac{DL} stand out as effective \ac{AI}-based  strategies for automating the differentiation of \acp{TN} in various contexts, including \ac{US}, fine-needle aspiration, and during thyroid surgical procedures \cite{ yang2019creating}.

Traditional methods for diagnosing \ac{TC}, like fine-needle aspiration biopsies, can often produce ambiguous outcomes, whereas \ac{AI} presents an opportunity for more accurate and less invasive alternatives. This review seeks to integrate insights from pathology, computer science, oncology, and radiology,  encouraging cross-disciplinary collaboration. It will also explore the clinical significance of \ac{AI}, offering recommendations for healthcare professionals on utilizing \ac{AI} advancements for improving patient care, and pinpointing directions for future research endeavours. Additionally, the review discusses the healthcare and economic system benefits, including cost savings and reduced wait times. Yet, it's essential to confront the challenges \ac{AI} brings, such as ethical considerations and data privacy, to facilitate its responsible integration into healthcare practices. This review aims to provide an extensive examination of \ac{AI}'s current and future impact on the detection of \ac{TC}, serving as a resource for both researchers and clinicians.

\subsection{Contribution of the paper}

This review explores the use of \ac{AI} in identifying \ac{TC}, emphasizing the shift towards improved diagnostic accuracy using \ac{AI} techniques in healthcare, specifically for detecting \ac{TC}. It begins with an overview of current frameworks and delves into \ac{AI} strategies such as \ac{SL}, \ac{USL} like clustering, and \acp{EM} like boosting and bagging, \ac{ML}, \ac{DL}, \acp{ViT}, \acp{LLM}. The significance of comprehensive datasets for \ac{AI} success is discussed, alongside an analysis of \ac{TCD}, feature selection, and extraction methods. It evaluates \ac{AI} effectiveness in  \ac{TCD} through various metrics and concludes by highlighting future research directions to overcome challenges and enhance \ac{AI} deployment in \ac{TCD}. The review underscores \ac{AI}'s potential to revolutionize \ac{TCD}, advocating for ongoing assessment to ensure ethical and effective use.\\

\noindent The main advancements presented in our paper are:
\begin{itemize}
\item A review of current frameworks coupled with a detailed investigation into diverse \ac{AI} strategies, including \ac{SL}, conventional classification, \ac{USL}, \ac{DL}, Transformers, and \acp{LLM} techniques.

\item An in-depth review of various \acp{TCD}, detailing their attributes and examining methods for feature selection and extraction used in different studies.

\item A detailed discussion on the benchmark criteria for assessing the efficacy of \ac{AI}-powered approaches in identifying \ac{TC}. These evaluation metrics cover a wide range, from regression and classification parameters to statistical, computer vision, and ranking parameters.

\item A thorough critique and exploration of the challenges, limitations, prevalent trends, and unresolved questions in the domain.

\item An analysis of future research priorities, highlighting specific areas that require further investigation to address current challenges and improve methods for detecting \ac{TC}.

\item A spotlight on the transformative impact of \ac{AI} in enhancing \ac{TC} diagnosis, stressing the importance of continual critical review to ensure its ethical and effective application.
\end{itemize}

\noindent Additionally, the main contributions of this review, as differentiated from other reviews, are summarized in Table \ref{table:1}.

\begin{table}[t!]
\leftskip=-4cm 
\caption{The notable advancements made by the suggested review in the categorization of \ac{TC} when contrasted with similar research endeavours. }
\label{table:1}
\scriptsize
\begin{tabular}{
m{2mm}
m{3mm}
m{8mm}
m{13mm}
m{4mm}
m{4mm}
m{4mm}
m{4mm}
m{4mm}
m{4mm}
m{5mm}
m{5mm}
m{4mm}
m{4mm}
m{4mm}
m{8mm}
m{8mm}
m{8mm}
}

\hline
Ref & Year & Patient& TC detect. & AI & ML & DL & Trans& LLM& F & TCE &  \multicolumn{6}{c}{Prospective path} & Metric \\

\cline{12-17}
 &  & privacy & schemes & apps &  &  & & & &  & IoMIT & RS & RL & PS &  XAI & EFC-AI  \\
\hline

\cite{liu2021deep} & 2021 & \cmark & \cmark & \xmark & \xmark & \xmark &  \cmark & \xmark & \cmark & \xmark & \xmark & \xmark & \xmark & \xmark & \xmark & \xmark & \xmark   \\

\cite{iesato2021role} & 2021 & \cmark & \cmark & \xmark  & \xmark & \xmark & \cmark & \xmark & \xmark & \xmark & \xmark & \xmark & \xmark  & \xmark & \xmark & \xmark & \xmark  \\

\cite{sharifi2021deep} & 2021 & \cmark & \cmark & \xmark  & \xmark & \cmark & \cmark & \xmark & \cmark& \cmark & \xmark & \xmark & \xmark & \xmark & \xmark & \cmark & \cmark   \\

\cite{lin2021deep} & 2021 & \cmark & \cmark & \xmark & \xmark  & \xmark & \cmark & \xmark &\xmark & \xmark & \xmark & \xmark & \xmark & \xmark & \xmark & \cmark & \cmark   \\

\cite{ha2021applications} & 2021 & \cmark & \cmark & \xmark  & \xmark & \cmark & \cmark & \xmark &\xmark &\xmark & \xmark & \xmark & \xmark & \xmark & \xmark & \cmark & \cmark   \\

\cite{wu2022deep} & 2022 & \cmark & \cmark & \xmark & \xmark & \xmark & \cmark & \xmark& \xmark& \xmark & \xmark & \xmark & \xmark & \xmark & \xmark & \xmark & \cmark  \\

\cite{pavithra2022deep} & 2022 & \cmark & \cmark & \xmark  & \xmark & \xmark & \cmark & \xmark&\xmark & \xmark & \xmark & \xmark & \xmark & \xmark & \cmark & \cmark & \xmark   \\

\cite{paul2022artificial} & 2022 & \cmark & \cmark & \xmark & \xmark & \cmark & \cmark & \xmark & \cmark& \xmark & \xmark & \xmark & \xmark & \xmark & \xmark & \cmark & \cmark   \\

\cite{ilyas2022deep} & 2022 & \cmark & \cmark & \xmark & \xmark & \cmark & \cmark & \xmark & \xmark & \xmark & \xmark & \xmark & \xmark & \xmark & \xmark & \cmark & \cmark   \\

\textbf{Our} & 2024 & \cmark & \cmark & \cmark & \cmark & \cmark & \cmark & \cmark & \cmark & \cmark & \cmark & \cmark & \cmark & \cmark & \cmark & \cmark & \cmark  \\
\hline
\end{tabular}
\begin{flushleft}
Abbreviations: 
Artificial intelligence applications (AI apps); Transformers (Trans.);
Features (F); \ac{TC} example (TCE); Internet of medical imaging thing (IoMT); Recommender systems (RS);  Reinforcement learning (RL); 
Panoptic segmentation (PS); Edge, fog and cloud networks based on AI (EFC-AI).  
\end{flushleft}
\end{table}

\subsection{Bibliometric analysis}
\begin{figure}[ht!]
\leftskip=-2cm 
\includegraphics[width=1\textwidth]{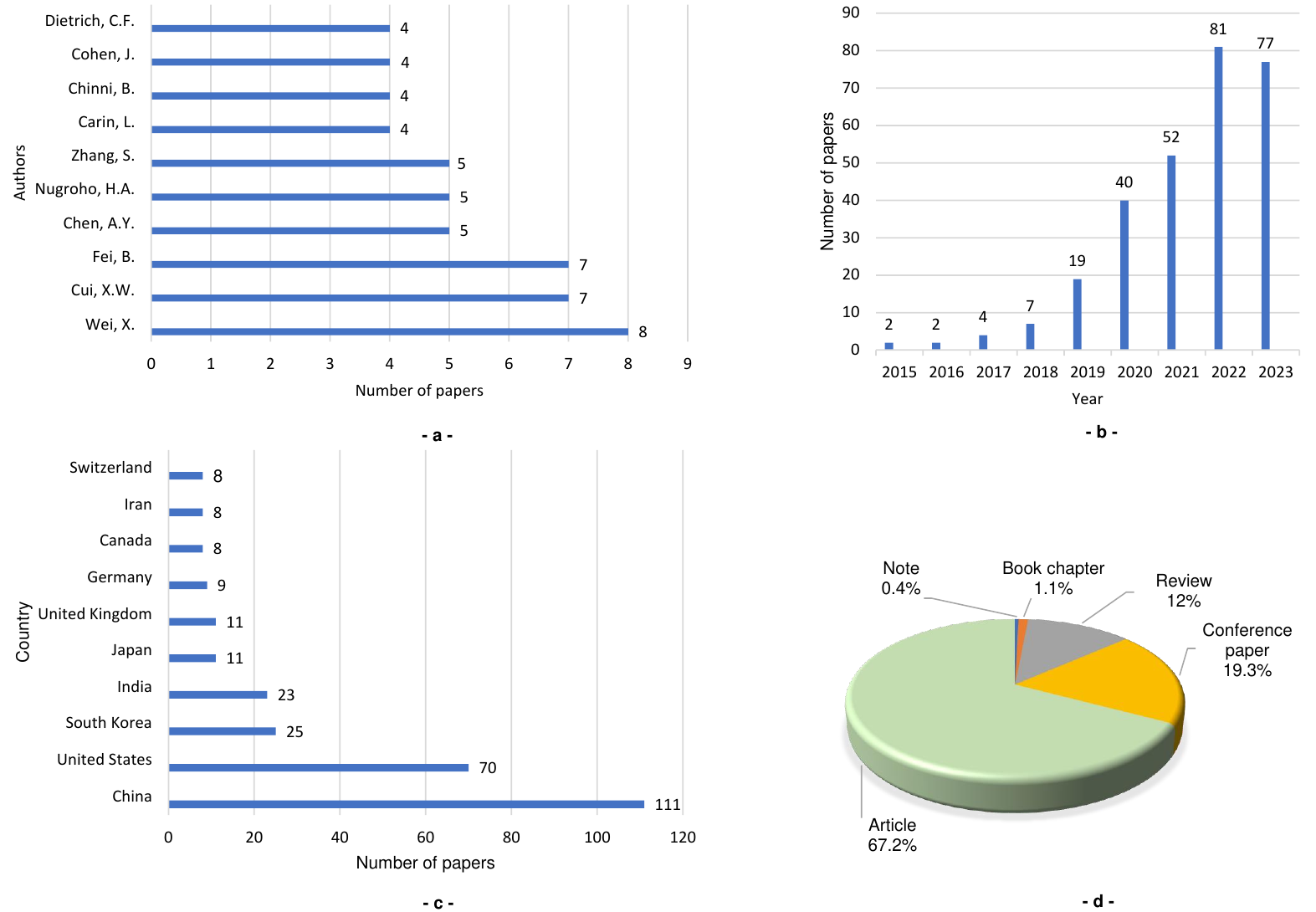}
\caption{Bibliometric analysis in terms of: (a) documents by author; (b) documents by year; (c) documents by country; (d) documents by type.}
\label{figbibio}
\end{figure}

A bibliometric analysis was performed to delve into and evaluate the scientific studies reviewed in this paper. The continuous interest in \ac{AI}-based \ac{TC} research is depicted in Figure \ref{figbibio}, with the publication count reaching 81 in 2022. Figure \ref{figbibio} (a) highlights the leading researchers in the field of TC-oriented \ac{AI} research, focusing on those who have published within the past five years. Figure \ref{figbibio} (b) provides a snapshot of the enduring interest in \ac{AI}-based \ac{TCD} research, showcasing a rising trend in the creation of \ac{AI} solutions for \ac{TC} prognosis and diagnosis since 2015. Figure \ref{figbibio} (c) maps out the countries that are major contributors to the research output in this area, with China and the United States showing a pronounced focus on \ac{AI}-driven \ac{TC} detection. Lastly, Figure \ref{figbibio} (d) illustrates the breakdown of publication types, with journal articles making up the bulk of the research (67.2\%), followed by conference papers (19.3\%).

\subsection{Roadmap}
The subsequent sections of this manuscript are organized as follows: Section \ref{sec2} discusses standardized assessment criteria and commonly used thyroid cancer datasets, emphasizing the importance of metrics and dataset features for AI-driven analysis in \ac{TCD}. Section \ref{sec3} provides a summary of current AI-based models and methods for \ac{TCD}, including classification, segmentation, and prediction frameworks. Section \ref{sec4} explores advanced \ac{TCD} methods leveraging \ac{ViT} and \ac{LLM}, focusing on their segmentation, classification, and prediction applications. Section \ref{sec5} outlines the limitations and challenges of existing AI methodologies, highlighting gaps in accuracy, data variability, and model generalizability. Section \ref{sec6} discusses future research directions, emphasizing innovative approaches to advance AI technologies in \ac{TCD}. Finally, Section \ref{sec7} presents the conclusion, synthesizing key insights and reaffirming the transformative potential of AI in medical diagnostics while offering perspectives for its integration into \ac{TCD} practices.

\section{Standardized assessment criteria and commonly used datasets}  \label{sec2}
\subsection{Metrics}

In this segment, we explore the standard metrics commonly utilized for evaluating \ac{TD} detection performance. These metrics act as critical benchmarks for measuring the success of methodologies, underscoring the significance of choosing the right metrics to assess \ac{ML} models. A variety of \ac{DL} metrics are used to determine the efficiency of the suggested method in identifying \ac{TD}. It is important to note that certain metrics have been previously addressed in \cite{kheddar2023deepSteg}. The other metrics, particularly designed for image processing applications in \ac{TD}, are concisely outlined in Table \ref{table:3}. 

\subsection{Datasets for TC}
In the context of \ac{TC} research, numerous datasets have been developed to support the testing and validation of \ac{ML} algorithms and models. This step is crucial, given the significant challenge of compiling these datasets within the endocrine \ac{ML} field. Table \ref{table:4} lists examples of publicly available \acp{TCD}.

{\scriptsize
\centering
\begin{longtable}[!ht]{m{0.5cm}m{2cm}m{4cm}m{6.5cm}}
\caption{Summary of the metrics for classification and regression employed in assessing \ac{AI}-driven methods for \ac{TCD}.}
\label{table:3}\\ 
\hline
&Metric & Mathematical formula & Description\\ \hline
\endfirsthead
\multicolumn{3}{c}{{Table \thetable\ (Continued)}} \\
\hline
&Metric & Mathematical formula & Description \\ \hline 
\endhead
\hline
\endfoot

\multirow{4}{*}{\rotatebox{90}{Classification and Regression }}& Specificity  & $\frac{T_{N}}{T_{N}+F_{P}}100\%$ &This metric represents the proportion of accurately predicted negative samples out of all the negative samples.  \\

& \Acf{RMSE}  & $\left( \sqrt{1-\left( ER\right) ^{2}}\right) \times SD$  & This is the standard deviation of the predicted errors between the training and testing datasets, and a lower value indicates the classifier's excellence.  \\

& \acs{JSI}  & $\frac{\left\vert A\cap B\right\vert }{\left\vert A\cup B\right\vert }= \frac{T_{P}}{T_{P}+F_{P}+F_{N}}$ & Paul Jaccard introduced this method to measure both the similarity and diversity among samples.  \\

& \Ac{VOE}  & $\frac{F_{P}+F_{N}}{T_{P}+F_{P}+F_{N}}$  & Assess the likeness between the segmented area and the ground truth area. \ac{VOE} quantifies the level of overlap between these two regions and is calculated as the ratio of the combined volume of the segmented and ground truth regions to the volume of their intersection.  \\

& \Ac{MAE}  & $\frac{1}{N}\sum\limits_{i=1}^{N}\left\vert a_{i}-p_{i}\right\vert$  & This measure indicates the average of the disparities between the real values and the predicted values.  \\[0.3cm]

\hline \\
& \Ac{SD}  & $\sqrt{\sum \left( x-\mu \right)^{2}/N}$  & It quantifies the degree of variability or spread within a dataset.  \\

\multirow{4}{*}{\rotatebox{90}{Statistical}} & \Ac{Corr}  & $\frac{(\sum ((x-\mu x)\cdot (y-\mu y)))}{(\sqrt{(\sum (x-\mu x)^{2})}\cdot \sqrt{(\sum (y-\mu y)^{2}))}}$  & It characterizes the extent of correlation or connection between two or more variables.   \\

& \acf{MRR}  & $\frac{1}{\left\vert Q\right\vert }\sum_{i=1}^{\left\vert Q\right\vert }%
\frac{1}{rank_{i}}$ &The \ac{MRR} is a statistical measure used to assess the average reciprocal rank of outcomes for a set of queries, as explained in \cite{lasseck2018audio}. Here, "rank$_{i}$" denotes the position at which the first relevant document appears for the i-th query.\\

& Kappa de Cohen & $k=\frac{\Pr \left( a\right) -\Pr \left( e\right) }{1-\Pr \left( e\right) }$  & This metric gauges the level of agreement between two assessors, considering chance as a baseline.   \\[0.3cm]

\hline

 & \Acf{PSNR}  & $10\cdot \log_{10}((MAX_{I}^{2})/MSE)$  & It quantifies the proportion between the highest achievable signal power and the power of the noise that impacts the faithfulness of its portrayal. \\

\multirow{9}{*}{\rotatebox{90}{Computer vision  
}} & Visual information fidelity (VIF)  & $\frac{\sum_{j}I(C^{j};F^{j}/s^{j})}{\sum_{j}I(C^{j};E^{j}/s^{j})}$  & It assesses the excellence of a reconstructed or compressed image or video in relation to the original signal. This evaluation considers how much visual information is retained in the processed image or video, accounting for the image's spatial and frequency attributes.  \\

& \Ac{NCC}  & $\frac{\sum_{i=1}^{M}\sum_{i=1}^{N}(I(i,j)-R(i,j))^{2}}{\sum_{i=1}^{M}\sum_{i=1}^{N}I(i,j)^{2}}$  & Assess the likeness between two images (or videos) by subtracting the mean value from each signal and subsequently normalizing the signals by dividing them by their standard deviation. Finally, compute the cross-correlation between the two normalized signals. \\

& \Ac{SC}  & $\frac{\sum_{i=1}^{M}\sum_{j=1}^{N}I(i,j)^{2}}{\sum_{i=1}^{M}\sum_{j=1}^{N}R(i,j)^{2}}$  & An elevated structural content value indicates that the image possesses lower quality.  \\

& \Ac{NVF}  & $\mathrm{Normalization}\left\{ \frac{1}{1+\delta_{bloc}^{2}}\right\}$    & This calculates the texture information within the image, where $\delta_{bloc}$ represents the variance in luminance. \\

& \Ac{VSNR}  & $10\log_{10}\left( \frac{C^{2}(I)}{(VD)^{2}}\right)$   & This approach sets distortion thresholds using contrast computations and wavelet transforms. \Ac{VSNR} is deemed excellent if distortions are below the threshold. It uses RMS contrast ($C(I)$) and visual distortion ($VD$).  \\

& \Ac{WSNR} & $10\log_{10}\left( \frac{\sum_{u=0}^{M-1}\sum_{v=0}^{N-1}\left\vert A(u,v) C(u,v) \right\vert ^{2}}{\sum_{u=0}^{M-1}\sum_{v=0}^{N-1}\left\vert A(u,v) -B(u,v) C(u,v) \right\vert ^{2}}\right)$   & It relies on the contrast sensitivity function, with $A(u,v)$, $B(u,v)$, and $C(u,v)$ denoting the 2D \ac{TFD}, as described in \cite{zhou2020weighted}.  \\

& \Ac{NAE}:  & $\frac{\sum_{i=1}^{M}\sum_{j=1}^{N}\left\vert I(i,j)-R(i,j) \right\vert }{\sum_{i=1}^{M}\sum_{j=1}^{N}\left\vert I(i,j) \right\vert }$  & This metric assesses the precision of an \ac{ML} model's predictions by quantifying the discrepancy between predicted and actual values relative to the range of actual values.  \\

& \Ac{LMSE}  & $\frac{\sum_{i=1}^{M}\sum_{j=1}^{N}\left[ L(I(i,j)) -L(R(i,j)) \right]^{2}}{\sum_{i=1}^{M}\sum_{j=1}^{N}\left[ L(I(i,j)) \right]^{2}}$    & It is a modified version of \ac{MSE}, utilizing the Laplacian distribution instead of the Gaussian distribution. $L(I(i,j))$ represents the Laplacian operator. \\[0.5cm]
\hline
\end{longtable}
\par}

\begin{scriptsize}
\begin{longtable}
{m{0.5cm}m{1cm}m{8cm}m{2.5cm}}
\caption{Instances of publicly available \acp{TCD} utilized in the identification of \ac{TC}.}
\label{table:4} \\
\hline
{Ref.} & {TCD} & {Description} & {Link} \\ \hline
\endfirsthead

\hline
{Ref.} & {TCD} & {Description} & {Link} \\ \hline
\endhead

\hline
\endfoot

\hline
\endlastfoot

{ \cite{d1}} & {THO} & This dataset is designed to investigate the fundamental causes and effects of \ac{TD} through the application of diverse omics approaches, including genomics, epigenomics, transcriptomics, proteomics, and metabolomics. & \href{https://transfer.sysepi.medizin.uni-greifswald.de/thyroidomics/}{Visit THO datasets} \\

{ \cite{d2}} & {TDD} & The dataset used for classification encompasses 5 features and 7200 instances, featuring a mix of 15 categorical and 6 numerical attributes. The classes within this dataset comprise hypothyroid, hyperfunction, and subnormal functioning. & \href{https://archive.ics.uci.edu/ml/datasets/thyroid+disease}{Visit TDS datasets} \\

\cite{d3} & {KEEL} & The KEEL dataset offers a collection of benchmarks for assessing the performance of different learning approaches, including semi-supervised classification and \ac{USL}. It encompasses 21 features, 7200 instances, and 3 classes. & \href{https://sci2s.ugr.es/keel/dataset.php?cod=67}{Visit KEEL datasets} \\

\cite{d4} & {GEO} & The \ac{GEO} database serves as a repository for genomics data. It is specifically structured to archive gene expression datasets, arrays, and sequences within \ac{GEO}. & \href{https://www.ncbi.nlm.nih.gov/geo/}{Visit GEO datasets} \\

{ \cite{d5}} & {DDTI} & The \ac{DDTI} dataset acts as an essential tool for both researchers and novice radiologists aiming to create algorithm-driven \ac{CAD} systems for analyzing \ac{TN}. It contains 99 cases and 134 images. & \href{http://cimalab.intec.co/?lang=en&mod=project&id=31}{Visit DDTI datasets} \\

{ \cite{d6}} & {NCDR} & The \ac{NCDR} functions as a repository for healthcare and research purposes, aimed at documenting every reported instance of cancer within England. The data originates from the Office for National Statistics. & \href{http://www.ncin.org.uk/about_ncin/}{Visit NCDR datasets} \\

{ \cite{d7}} & {PLCO} & The National Cancer Institute backs the \ac{PLCO} cancer screening trial, which focuses on identifying the primary factors influencing cancer incidence in both genders. This trial encompasses records from 155,000 participants and includes comprehensive studies on \ac{TC} incidence and mortality. & \href{https://prevention.cancer.gov/major-programs/prostate-lung-colorectal-and-ovarian-cancer-screening-trial}{Visit PLCO datasets} \\
\end{longtable}
\end{scriptsize}

\section{Summary of current models and methods} \label{sec3}

\begin{figure}[t!]
\centering
\includegraphics[width=0.55\textwidth]{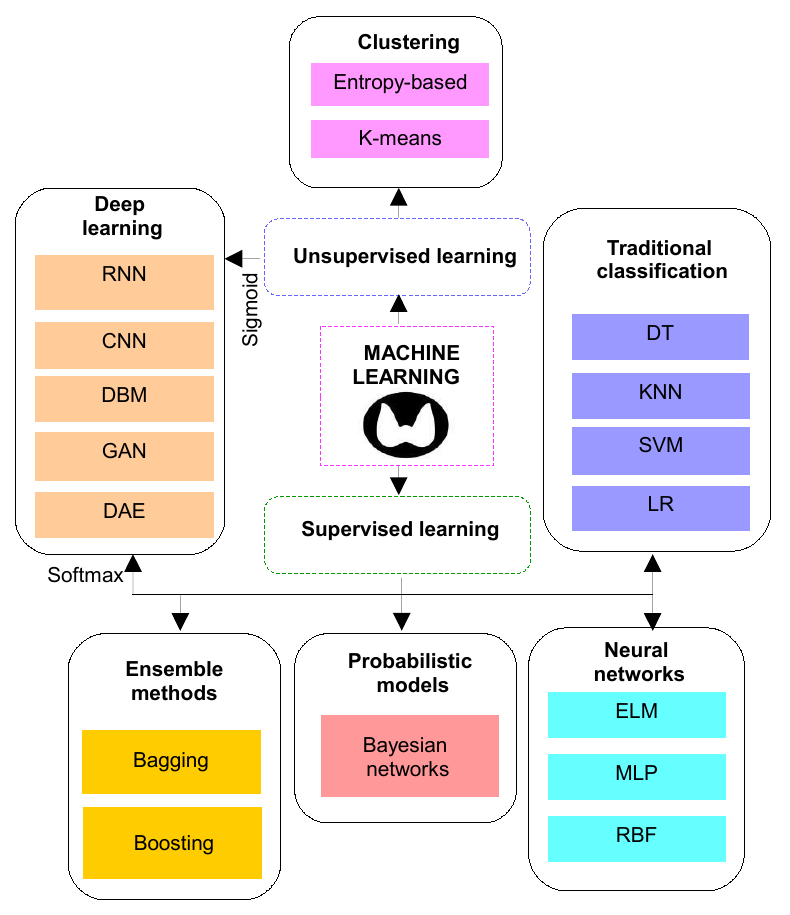}
\caption{Classification of \ac{TCD} strategies utilizing \ac{AI}.} 
\label{fig3}
\end{figure}

This section discusses the various \ac{AI}-based methodologies utilized for diagnosing \ac{TG} cancers. Figure \ref{fig3} offers a visual depiction of the classification system for \ac{TC} diagnosis methods leveraging \ac{AI}.  

\subsection{Purpose of AI-driven examination}
This review centres on the use of \ac{AI} for identifying \ac{TC}. Comprehending the foundational objectives of each framework is essential for acquiring a more profound understanding of their reasoning \cite{habchi2023ai}.

\noindent \textbf{(a) Classification of thyroid carcinoma}: Entails the sorting of TCs according to their histopathological features, clinical manifestations, and prognostic outcomes. Various forms of thyroid carcinomas exist, each defined by unique characteristics. The principal categories are: (i) \ac{PTC}: Representing the most prevalent form, \ac{PTC} constitutes approximately 80\% of all \ac{TC} cases. It typically exhibits slow growth but has a propensity to metastasize to neck lymph nodes. Nonetheless, \ac{PTC} generally responds well to treatment. (ii) \ac{FTC}: Ranking as the second most frequent type, \ac{FTC} has the capability to invade blood vessels and spread to distant body parts, although it is less prone to lymph node metastasis. (iii) \ac{MTC}: Arising from the parafollicular or C cells of the thyroid, which secrete calcitonin, an increase in blood calcitonin levels may signal \ac{MTC}. (iv) \ac{ATC}: \ac{ATC} is a highly aggressive and rare \ac{TC} variant, characterized by its rapid spread to other neck regions and the body, making it challenging to treat. The stratification of thyroid carcinomas is vital for selecting the optimal treatment plan for individual patients, considering tumour dimensions, location, patient age, and general health. The evolution of \ac{AI} and \ac{ML} has significantly contributed to the automation and enhancement of thyroid carcinoma classification accuracy. Various models have been devised to categorize tumours based on medical imaging or genetic information. For example, Liu and colleagues \cite{liu2018setsvm} highlight the fundamental importance of \ac{SVM} in the detection of cancer. In a similar vein, Zhang and their team \cite{zhang2019integrating, zhang2019diagnosis} introduce approaches utilizing \ac{DNN} to distinguish between malignant and benign  \acp{TN} in \ac{US} imagery. Moreover, the \ac{Bi-LSTM} model \cite{chen2018thyroid}, shows noteworthy precision in the classification of \acp{TN}. These classification approaches create structured hierarchies crucial for organizing knowledge and processes in the field of \ac{TCD}.

\noindent \textbf{(b) Segmentation of thyroid carcinoma}: Segmentation plays a pivotal role in the detection of \ac{TC} by enabling the accurate isolation and analysis of the thyroid gland, along with any potentially suspicious nodules or lesions present \cite{ma2023amseg}. Often detected through medical imaging modalities like \ac{US}, \ac{CT} scans, or \ac{MRI}, \ac{TCD} requires the precise delineation of the \ac{ROI} for sound diagnostic judgements. Segmentation not only facilitates the distinction of the thyroid gland from the surrounding tissues but also supports the precise quantification of nodule dimensions and volume, which are critical for evaluating the potential for malignancy. Furthermore, it enables the extraction of significant image attributes such as texture and shape, providing critical data for \ac{ML} models or other analytical methods to improve diagnostic precision. Segmented images also enhance visual clarity, aiding radiologists and medical practitioners in the visual assessment and interpretation of areas of concern within the thyroid, vital for detecting abnormalities indicative of cancer. For longitudinal analyses, segmentation is invaluable in tracking changes in the thyroid and nodules over time, monitoring disease evolution or the efficacy of treatments. It also plays a role in accurately locating biopsy sites for suspicious nodules, guaranteeing targeted sample collection for cancer verification. In the context of treatment planning, segmentation is instrumental in assessing the tumour's size and its relationship to vital anatomical structures, thereby guiding therapeutic decisions. Moreover, the introduction of automated segmentation technologies streamlines clinical workflows by minimizing manual input and variability, empowering medical experts to dedicate more attention to complex diagnostic activities. Consequently, the segmentation process in \ac{TCD} enhances precision, consistency, and confidence in diagnostics, markedly impacting patient management and outcomes. \ac{AI} methods, especially \ac{CNN} and the U-Net architecture, are becoming progressively popular for the segmentation of thyroid carcinoma. Their growing preference is largely due to their capacity to learn from and generalize across large datasets, significantly improving the accuracy and dependability of the segmentation procedure.

\noindent \textbf{(c) Prediction of thyroid carcinoma}:  Prediction in \ac{TCD} involves utilizing diagnostic tools and \ac{ML} models to estimate the risk of development based on factors like genetic predisposition, gender, age, radiation exposure, and lifestyle choices. It's important to note that predictions indicate a heightened risk rather than a definite outcome. Medical practices often combine various predictive assessments to improve accuracy. For example, \ac{ML} algorithms developed from medical records can help distinguish between benign and malignant nodules, facilitating early intervention. Studies, such as one utilizing \ac{ANN}  and \ac{LR}, and another employing a \ac{CNN} to analyze over 10,000 microscopic \ac{TCD} (images) \cite{sajeev2020thyroid}, demonstrate the application of predictive techniques in identifying \ac{TC} risk, showcasing advancements in \ac{AI}-driven predictive modelling for more effective treatment strategies.

\subsection{Pre-processing}

\Ac{PCA} operates as a sophisticated method for preprocessing, transforming samples (variables) into a smaller set of uncorrelated ones. This technique effectively reduces the volume of variables, thus cutting down on redundant data, all the while striving to maintain the integrity of the data relationships. \ac{PCA} is extensively used in the realms of cancer detection and distinguishing between malignant and benign thyroid cells. In research conducted by Shankarlal et al. \cite{shankarlal2020performance}, \ac{PCA} was deployed to filter the most relevant set of wavelet coefficients from \ac{DTCW} processed noisy thyroid images, which were then categorized using \ac{RF}. In another instance, Soulaymani et al. \cite{soulaymani2018epidemiological} applied \ac{PCA} to a dataset comprising 399 patients with three different \ac{TC}, allowing the classification based on variables like age, sex, type of cancer, and geographical location.

\subsection{ML features selection and extraction }

\subsubsection{Selection methods}
 The primary goal of the selection process is to identify and select relevant features that can enhance the accuracy of classification, simultaneously eliminating non-essential variables \cite{cui2018ovarian}. Many techniques of feature selection are proposed:

\noindent \textbf{(a)  \Acf{CFS}:}
The \ac{CFS} technique is commonly applied to examine the relationships between various cancer-related attributes. Although \ac{CFS} offers a range of advantages, there are also notable limitations associated with its use in feature selection within \ac{ML}:

\begin{itemize}
    \item \textbf{Advantages:}  \ac{CFS} is valued for its straightforwardness, ability to identify linear correlations, capability to decrease dimensionality, and potential to boost model efficiency and clarity. This method supports quicker model training and prediction, exhibits robustness against outliers, and accommodates the incorporation of expert insights. Moreover, it enhances the effectiveness of other methodologies, provides opportunities for visualization and deeper understanding, lowers expenses, and enables the conduct of sensitivity analyses.
    
    \item \textbf{Disadvantages:} It's important to recognize that this approach may miss complex, nonlinear relationships between features and the outcome variable. Additionally, it can be vulnerable to multicollinearity, where features are highly correlated with each other, requiring additional preprocessing steps. Careful consideration of the specific issue and dataset at hand is crucial when applying this method.
\end{itemize}

The \ac{CFS} algorithm has been widely incorporated into feature selection strategies to enhance classification outcomes across various studies. For example, in \cite{al2019gene}, researchers utilized \ac{CFS} for feature selection within microarray datasets, successfully minimizing data dimensionality and pinpointing significant genes. A combined model that blended \ac{CFS} with binary particle swarm optimisation was developed in \cite{jain2018correlation} for cancer classification, and applied to 11 standard microarray datasets. Additionally, the CSVM-RFE technique, which integrates \ac{CFS}, was applied in \cite{rustam2018correlated} to diminish the feature set in cancer research by removing non-essential elements. Moreover, \ac{CFS} methodologies were utilized in \cite{bhalla2020expression} for the identification of key RNA expression features.\\

\noindent \textbf{(b) \Acl{RA}:}
The \ac{RA} is an effective technique used in feature selection, evaluating the discriminative power of features between classes through score assignment. \ac{RA} has its own array of advantages and disadvantages:

\begin{itemize}
    \item \textbf{Advantages:} \ac{RA} offers several benefits in feature selection, including its robustness against noisy data, capability to handle both continuous and categorical features, and ability to detect feature interactions without assuming their independence. It also reduces bias in datasets with imbalanced classes, eliminates the need for model training, and facilitates sensitivity analysis. These attributes make \ac{RA} an advantageous tool for feature selection in various data scenarios.
    
    \item \textbf{Disadvantages:}  \ac{RA} exhibits significant computational complexity, affecting its applicability to large datasets. Its performance is sensitive to parameters, particularly the choice of the number of nearest neighbours (k), which can be challenging to optimize. The stability of \ac{RA} is also affected, with variations in the dataset leading to different selections of features. Furthermore, it is designed solely for use within \ac{SL} contexts, struggles with non-metric features, and necessitates adaptations for handling multiclass classification scenarios.
    
\end{itemize}

This method evaluates the importance of different features by exploring the relationships between variables related to cancer. In their research, Cui et al. \cite{cui2018ovarian} suggested a feature selection strategy that employs the \ac{RA} algorithm to enhance its effectiveness.

\subsubsection{Extraction methods}

\noindent \textbf{(a) \acs{PCA}:} \Acf{PCA} has been widely recognized in numerous pieces of research for its effectiveness in reducing data dimensionality and decoupling cancer-related features. \ac{PCA} is praised for its ability to decrease dimensions and reveal patterns, although it may compromise on interpretability and is optimally used with linear correlations. For example, Shankarlal et al. (2020) implemented \ac{PCA} to refine feature selection for \ac{TCD} via the \ac{DTCW} transformation \cite{shankarlal2020performance}. Soulaymani et al. (2018) investigated \ac{PCA}'s capability in distinguishing various \ac{TC} subtypes, such as papillary, follicular, and undifferentiated types \cite{soulaymani2018epidemiological}. Additionally, O et al. (2019) assessed \ac{PCA} and linear discriminant analysis in the classification of Raman spectra for different \ac{TC} subtypes \cite{o2019raman}. 

\noindent \textbf{(b)  Texture description:}
Texture analysis is a highly regarded technique for extracting related data in \ac{TC} segmentation, classification, and prognosis efforts. The scientific community has developed various texture analysis methods, including wavelet transforms, binary descriptors, and statistical descriptors, among others. Specifically, the \ac{DWT} has garnered significant interest for its exceptional capability in data decorrelation. Although texture analysis is beneficial for distinguishing textures, it can be affected by changes in lighting conditions and does not inherently understand semantic content, which may limit its application in complex visual tasks. Wavelet-based methods have been extensively applied in detecting \ac{TC}. For example, Sudarshan et al. \cite{sudarshan2016application} applied wavelet techniques to identify cancerous areas in thyroid, breast, ovarian, and prostate tumours. Additionally, Haji et al. \cite{haji2019novel} used texture data for the diagnosis of \ac{TN} malignancy employing a 2-level 2D wavelet transform. Further contributions to this field are documented in studies such as \cite{yu2019transverse} and \cite{nguyen2019artificial}.

\noindent \textbf{(c) \Acf{AC}:} The \ac{AC} model, a versatile framework often used in image processing, was initially introduced by Kass and Witkin in 1987. \ac{AC} is known for its ability to adjust to complex shapes, yet it faces challenges such as sensitivity to initial placements and issues with overlapping figures. Various strategies have been developed to address these challenges in contour segmentation using deformable curve models. These models have seen significant application in \ac{TCD}, as evidenced by research conducted by \cite{poudel2017active}, \cite{poudel20163d}, and \cite{nugroho2015thyroid}.

\noindent \textbf{(d) LBP and GLCM:}
\Ac{LBP} are descriptors used in computer vision for identifying textures or objects in digital images. They are appreciated for their straightforwardness and ability to distinguish features effectively. However, \ac{LBP} may be vulnerable to noise and often requires tuning of parameters for optimal performance. The \ac{LBP} method was employed in \ac{TCD}, as illustrated in a study by Yu et al. \cite{yu2019transverse}. Furthermore, the integration of \ac{LBP} with \ac{DL} has been explored for distinguishing between benign and malignant \ac{TN}, as seen in the studies by Xie et al. \cite{xie2020hybrid} and Mei et al. \cite{mei2017thyroid}.

The \ac{GLCM} serves as a tool to depict the occurrence frequency of pixel value pairs at a predetermined distance within an image. It is particularly useful for texture analysis and the identification of distinctive features. Nevertheless, \ac{GLCM} faces challenges such as sensitivity to image variations, high computational demands, and the need for careful parameter tuning. For example,  Dinvcic et al. \cite{dinvcic2020fractal} employed \ac{GLCM} in a comparative study to investigate the differences between patients with Hashimoto's thyroiditis-associated \ac{PTC} and those with Hashimoto's thyroiditis only.

\noindent \textbf{(e) ICA:} In \ac{ICA}, data is decomposed into a set of independent contributing features to aid in feature extraction. \ac{ICA} is adept at identifying statistically independent components, making it valuable for tasks like source separation. Its strengths lie in uncovering non-linear relationships and facilitating data compression. Nonetheless, \ac{ICA} faces challenges due to assumptions about the data mixing process and can be difficult to interpret. \ac{ICA} is applied for disentangling multivariate signals into their separate constituents. In the research conducted by Kalaimani et al. \cite{kalaimani2019analysis}, \ac{ICA} was used to isolate 29 attributes as independent and significant features for categorizing data into hypothyroid or hyperthyroid groups through a \ac{SVM}. A portrayal of the techniques derived from \ac{ML}/\ac{DL} utilized in diagnosing \ac{TC} is provided in Table \ref{table:6}.

\begin{table}[h!]
\leftskip=-4cm 
\caption{Summary of features methods based on \ac{ML}/\ac{DL} conducted in the diagnosis of \ac{TC}.}
\label{table:6}
\scriptsize

\renewcommand{\arraystretch}{1.5} 
\begin{tabular}{
m{5mm}
m{10mm}
m{10mm}
m{10mm}
m{14mm}
m{110mm}
}
\hline

{\small Ref.} & {\small Year} & \ac{ML}/\ac{DL} & {\small Classifier} & {\small Features} & {\small Contributions}\\ \hline

{\small \cite{ahmad2017thyroid}} & 2017 & ML & KNN & FC/IG &  Minimize data duplication and decrease processing duration. KNN addresses absent dataset values, while ANFIS receives the modified data as input. \\

{\small \cite{nugroho2017classification}} & 2017 & \ac{ML} & \ac{SVM} & FC/CFS &  Retrieve the geometric and moment characteristics, while specific \ac{SVM} classifier kernels categorize the acquired features.\\

{\small \cite{mourad2020machine}} & 2020 & DL& CNN & FC/R &  Utilize both \ac{ML} techniques and feature selection algorithms, specifically Fisher's discriminant ratio, Kruskal-Wallis analysis, and Relief-F, for the examination of the SEER database. \\

{\small \cite{song2022rapid}} & 2022 & DL& CNN & FE/\ac{PCA} & This study mitigated the impact of imbalanced serum Raman data on prediction outcomes by employing an oversampling technique. Subsequently, the dimension of the data was reduced with \ac{PCA} before applying \ac{RF} and the Adaptive Boosting for classification.\\

{\small \cite{acharya2012thyroscreen}} & 2012 & ML & Boosting & FE/TD & Integrate \ac{CAD} with \ac{DWT} and extract texture features. Utilize the AdaBoost classifier to classify images into either malignant or benign thyroid images based on the extracted features.\\

{\small \cite{nugroho2021computer}} & 2021 &DL & CNN & FE/AC & Improve image quality, perform segmentation and extract multiple features, including both geometric and texture features. Each feature set is subsequently classified using MLP and \ac{SVM}, leading to the classification of either malignant or benign cases. \\

{\small \cite{sun2020evaluation}} & 2020 & ML & \ac{SVM} & FE/LBP & Deep features are obtained through \ac{CNN}, and they are merged with manually crafted features, which include \ac{HOG} and scale-invariant feature transforms, to generate combined features. These combined features are subsequently employed for classification via an \ac{SVM}. \\

{\small \cite{liu2019value}} & 2019 & ML & \ac{SVM} & FE/GLCM &  Apply a median filter to mitigate noise and outline the contours before feature extraction from thyroid regions, encompassing \ac{GLCM} texture features. Subsequently, employ \ac{SVM}, \ac{RF}, and Bootstrap Aggregating (Bagging) to differentiate between benign and malignant nodules. \\

{\small \cite{kalaimani2019analysis}} & 2019 & ML& \ac{SVM} & FE/ICA & A multi-kernel-based classifier is employed for thyroid disease classification.  \\
\hline
\end{tabular}
\end{table}

\subsection{SL-based TCD classification}

\Ac{SL} provides high accuracy, interpretability, and robust predictive capabilities. However, it necessitates extensive labeled data, poses risks of overfitting, and can be computationally intensive \cite{liu2023self}. The following algorithms represent prominent \ac{SL} techniques utilized in \ac{TCD}:

\color{black}

\noindent \textbf{(a)  DT and LR:}  \Ac{DT} learning is a technique in data mining that uses a model for predictive decision-making. In such a model, the outcomes are indicated by the leaves, and the branches represent the input features. This method has been utilized in detecting latent thyroid disorders, as evidenced by a range of studies, such as in \cite{yadav2020prediction}. In the research presented in \cite{zhao2015logistic}, \ac{LR} was employed to pinpoint specific characteristics of thyroid microcarcinoma among a group of 63 patients. This analysis utilized data from both \ac{CEUS} and traditional \ac{US} evaluations. Furthermore, a significant study from northern Iran, detailed in \cite{yazdani2018factors}, used \ac{LR} to investigate a large dataset encompassing 33,530 cases of \ac{TCD}. \ac{LR} is a widely used binomial regression model within the domain of \ac{ML}.
\vskip2mm

\noindent \textbf{(b) ELM and MLP:} The \ac{ELM} model is distinguished by a single layer of hidden nodes that possess randomly assigned weight distributions. Crucially, the process of determining weights between the inputs and the hidden nodes to the outputs is executed in a solitary step, rendering the learning mechanism markedly more efficient than that of alternative models. The efficacy of the \ac{ELM} approach in diagnosing \ac{TD} has been corroborated through various research efforts \cite{pavithra2021optimal}.

The \ac{MLP} is a type of feed-forward network that directs data processing sequentially from the initial point to the final layer of output. Within this architecture, each layer is made up of a different number of neurons. Rao and colleagues \cite{rao2019thyroid} devised an innovative approach for categorizing \acp{TN} employing an \ac{MLP} integrated with a backpropagation learning mechanism. Their design comprised four neurons in the initial layer, three neurons in each of its ten concealed layers, and one neuron in the terminal layer. In a separate effort to enhance the precision of \ac{TD} diagnosis, Hosseinzadeh et al. \cite{hosseinzadeh2020multiple} utilized \ac{MLP} networks. Their analysis compared the efficacy of \ac{MLP} networks against the backdrop of existing research on \ac{TCD} classification, highlighting the superior performance of \ac{MLP} networks. 

\vskip2mm

\noindent \textbf{(c) PM and EM: } \acp{PM}, such as Bayesian networks, are vital in computer science and statistics for modeling uncertainties and variable dependencies. They support decision-making in \ac{ML}, data analysis, and parameter estimation. Bayesian networks use directed acyclic graphs, aiding in reasoning under uncertainty, predictions, and medical diagnosis of \acp{TN}.  In the realm of oncology research, tackling the intricacies of cancer datasets and enhancing the accuracy of detection frequently involves the use of \acp{EM}. This strategy splits the dataset into several subsets, upon which a variety of \ac{ML} algorithms are applied in parallel. The insights gained from these individual algorithms are subsequently merged to derive a comprehensive diagnosis. The main goal behind adopting \acp{EM} is to forge a superior predictive model tailored for the detection of \ac{TC}. Such an approach has been validated in multiple studies, including a significant one conducted by Chandran et al. \cite{chandran2021diagnosis}, where the authors underscored the contribution of \acp{EM} to a more profound data comprehension and heightened diagnostic accuracy.

\vskip2mm

\noindent \textbf{(d) Bagging and Boosting: } Bagging is a notable ensemble learning approach in \ac{TCD}, aimed at boosting the accuracy and consistency of \ac{ML} algorithms. This technique achieves its objectives by lowering variance and offering protection against overfitting. It finds broad application in a variety of methods, with a particular emphasis on \ac{DT}. The main goal of Bagging is to improve the effectiveness of weaker classifiers in the context of \ac{TCD} screening. In their research, Chen et al. \cite{chen2020diagnosis} presented \ac{FB} as an ensemble learning strategy designed to reduce the correlation between models in an ensemble. \ac{FB} accomplishes this by training each model on randomly selected feature subsets from the dataset, rather than using the full set of features. The utility of \ac{FB} is demonstrated in its ability to distinguish between benign and malignant cases of \ac{TC} \cite{himeur2021artificial}. Within the scope of \ac{USL}, meta-algorithms play a crucial role in reducing variance and improving the performance of weak classifiers, effectively converting them into robust classifiers \cite{mehta2019high}.

In the context of boosting, Pan et al. in their study \cite{pan2016improved} employed a novel method called AdaBoost to identify \ac{TN}, utilizing the widely recognized \ac{UCI} dataset. The classification was performed using the random forest method, with \ac{PCA} employed to retain data variance. Chen et al. \cite{chen2016xgboost}, the \ac{XGBoost} algorithm was highlighted as a powerful implementation of gradient-boosted \ac{DT}, with its application extending across multiple research areas including sports and health monitoring \cite{guo2019xgboost}. Specifically, in the context of \ac{TC}, the \ac{XGBoost} algorithm was employed by researchers to distinguish between benign and malignant \ac{TN} \cite{chen2020computer}, offering a solution to the problem of obtaining accurate diagnoses without the need for large datasets that \ac{DL} models usually require.

\color{black}
\subsection{USL-based TCD classification}
\Ac{USL} is the process of analyzing data that hasn't been previously labelled or annotated. Its primary goal is to uncover the underlying structures within datasets that do not have predefined labels. Contrary to \ac{SL}, which depends on labelled data for evaluating its effectiveness, \ac{USL} operates without such direct guidance, presenting additional challenges in result assessment. Although \ac{USL} algorithms are capable of addressing more complex problems than their supervised counterparts, they might also lead to increased uncertainty, sometimes creating unintended categories or incorporating noise rather than identifying clear patterns. Nonetheless, \ac{USL} is considered an indispensable asset in \ac{AI}, offering the potential to detect patterns within data that may not be obvious at first \cite{nobile2023unsupervised}.

Clustering is one of the important techniques in \ac{USL}. The objective of this strategy is to organize \ac{TCD} into distinct, uniform groups that share similar features. This process aids in the categorization of unlabeled data into malignant or benign sections. Due to its straightforwardness, this technique has received significant attention in numerous medical research areas, enhancing its applicability to tasks like identifying breast cancer \cite{agrawal2019combining}, and discovering brain tumours \cite{khan2021brain}. Clustering methods also prove helpful in classifying cancer instances that are not clearly defined \cite{yu2017clustering}. A research documented in \cite{chandel2020analysing} employed clustering to determine factors impacting the normal functioning of the thyroid gland. The use of \ac{PCA} played a key role in organizing the clusters and simplifying the data structure. Additionally, an innovative automated clustering system for diagnosing \ac{TC} was developed, as described in \cite{katikireddy2020performa}, which recommended appropriate medication treatments for hyperthyroidism, hypothyroidism, and normal cases. As an example, the study in \cite{venkataramana2018comparative} explored the use of fuzzy clustering on thyroid and liver datasets from the \Ac{UCI} repository, where \ac{FCM} and \ac{PFCM} algorithms were employed and their performances compared.

\vskip2mm

\noindent \textbf{(a) \Ac{KM}:} The \ac{KM} method is used for dividing data into partitions and tackles a combinatorial optimization challenge. It is commonly used in \ac{USL}, categorizing observations into k distinct clusters. In the research presented by Mahurkar et al. \cite{mahurkar2017normalization}, the study investigates the application of \ac{ANN} and an improved K-Means algorithm to standardize raw data. This study employed a thyroid dataset from the \ac{UCI} repository, comprising 215 total instances.

\vskip2mm

\noindent \textbf{(b) Entropy-based (EB):} In the study conducted by Yang et al. \cite{yang2019information}, a novel, parameter-free computational model called DeMine was introduced for the prediction of \acp{MRM}. DeMine utilizes an information entropy-based methodology, comprising three primary steps. The process begins by converting the miRNA regulation network into a cooperative \acp{MRM} network. It then proceeds to pinpoint miRNA clusters, aiming to maximize entropy density within the specified cluster. The final step involves grouping co-regulated miRNAs into their appropriate clusters, thereby finalizing the \acp{MRM}. This technique enhances predictive precision and facilitates the identification of a broader array of miRNAs, potentially acting as tumour markers in cancer diagnosis.

\subsection{DL-based TCD  classification}

\Ac{DL} surpasses traditional \ac{ML} by automating feature extraction, enhancing performance with large datasets, and offering scalability for complex models.  Table \ref{tab:dl_ml_comp} presents a comparison between \ac{ML} and \ac{DL} in the context of \ac{TCD}. Key methods for \ac{TCD} include \acp{CNN} for image analysis, \acp{RNN} for sequential data, \ac{TL} for leveraging pre-trained models \cite{kheddar2023deepIDS}, ensemble learning for robustness, and attention mechanisms for focused detection. These advancements enable more accurate and efficient early detection, improving patient outcomes.

\begin{table}[ht!]
\leftskip=-4cm 
    \caption{Comparison of  \ac{ML} and \ac{DL} for \ac{TCD}.}
    \label{tab:dl_ml_comp}
    \scriptsize
    \begin{tabular}{l|p{6.5cm}|p{6.5cm}}
        \hline
        \textbf{Criteria} & \textbf{\ac{DL} } & \textbf{\ac{ML}} \\
        \hline
        \textbf{Best Suitable Scenario} & Large datasets with complex features relevant to thyroid cancer characteristics & Smaller datasets with simpler features, suitable for initial screening \\
        \hline
        \textbf{Advantages} & 
        \begin{itemize}
            \item High accuracy with large, diverse data
            \item Automated feature extraction from complex medical images and data
            \item Capable of identifying subtle patterns indicative of thyroid cancer
        \end{itemize} &
        \begin{itemize}
            \item Effective with smaller datasets, reducing the need for extensive data collection
            \item Easier interpretation and understanding of models by medical professionals
            \item Faster training times suitable for rapid decision-making in clinical settings
        \end{itemize} \\
        \hline
        \textbf{Disadvantages} & 
        \begin{itemize}
            \item Requires large labelled datasets for training, which can be challenging to acquire and annotate
            \item Computationally intensive, requiring high-performance hardware and longer processing times
        \end{itemize} &
        \begin{itemize}
            \item Requires manual feature engineering to extract relevant medical features
            \item Limited performance in capturing intricate patterns or anomalies in complex thyroid conditions
        \end{itemize} \\
        \hline
        \textbf{Performance} & Excels in analyzing high-dimensional medical imaging data & Performs well with structured clinical data from standard medical tests \\
        \hline
        \textbf{Resource Requirements} & High (GPUs, memory, computational power) due to complex data processing & Moderate (CPUs, less memory) sufficient for structured data analysis \\
        \hline
        \textbf{Risk of Overfitting} & High, mitigated with regularization techniques and extensive validation & Moderate, mitigated with cross-validation strategies \\
        \hline
        \textbf{Future Development} & Integration with advanced \ac{USL} for discovering new disease markers and improved annotation techniques & Enhanced algorithms for feature selection and hybrid models combining DL's imaging capabilities with ML's interpretability \\
        \hline
    \end{tabular}
\end{table}

\color{black}

Typically, the network's depth facilitates the extraction of increasingly abstract and advanced features as data moves through its layers. By leveraging large neural networks with several layers, \ac{DL} can independently learn, create, and improve data representations, which is why it is known as "deep" learning. Within the realm of \ac{TCD} , \ac{DL} is instrumental in several areas, including: (i) Classification of image: \ac{DL} techniques, for example \ac{CNN}, are trained to categorize thyroid \ac{US} images, distinguishing between malignant and benign nodules by analyzing texture and shape, and other features \cite{canton2021automatic, peng2021deep}, streamlining the process and assisting with the early identification of \ac{TC}; (ii) Analysis of disease: \ac{DL} is used to examine cytopathological or histopathological slide images, aiding in identifying and classifying cells of cancerous; (iii) Analysis of genomic information: In this era \ac{DL} models are capable of analyzing genetic variations linked to the risk of \ac{TC}; (iv) Radiomics: \ac{DL} models are adept at extracting multidimensional information obtained from radiographic images, contributing to more accurate and individualized treatment strategies; and (v) Predictive analysis: By analyzing electronic health records and patient information, \ac{DL} models can forecast the probability of an individual developing \ac{TC}, facilitating early intervention.  The following \ac{DL} algorithms are the widely used techniques for \ac{TCD}:

\vskip2mm

\noindent \textbf{(a) DAE:} \Acp{DAE} play a pivotal role in the identification of \ac{TC} by adeptly deriving significant features from \ac{US} or histopathological imagery. As a subset of \ac{ANN}, \ac{DAE} are primarily focused on the accurate reconstruction of inputted data. Their usage extends to tasks such as reducing dimensions and enhancing feature learning capabilities. The integration of \acp{DAE} into the workflow for classifying thyroid carcinoma typically unfolds in several phases: (i) initial data pre-processing, (ii) creation of perturbed input data, (iii) training the \ac{DAE}, (iv) extraction of relevant features, and (v) execution of the classification procedure. In the research presented by Ferreira et al. \cite{ferreira2018autoencoders}, a variety of six autoencoder models were utilized for the purpose of classifying \ac{PTC}, incorporating strategies such as the stabilization of weights and network fine-tuning. The architecture of these autoencoders, especially the encoding layers, played an integral role in the effective integration of the network. In a related study by Teixeira et al. \cite{teixeira2017learning}, both \acp{DAE} and their stacked configurations were applied to distill crucial features and pinpoint genes relevant to \ac{TC} diagnosis.

\begin{figure}[t!]
\centering
\includegraphics[scale=0.75]{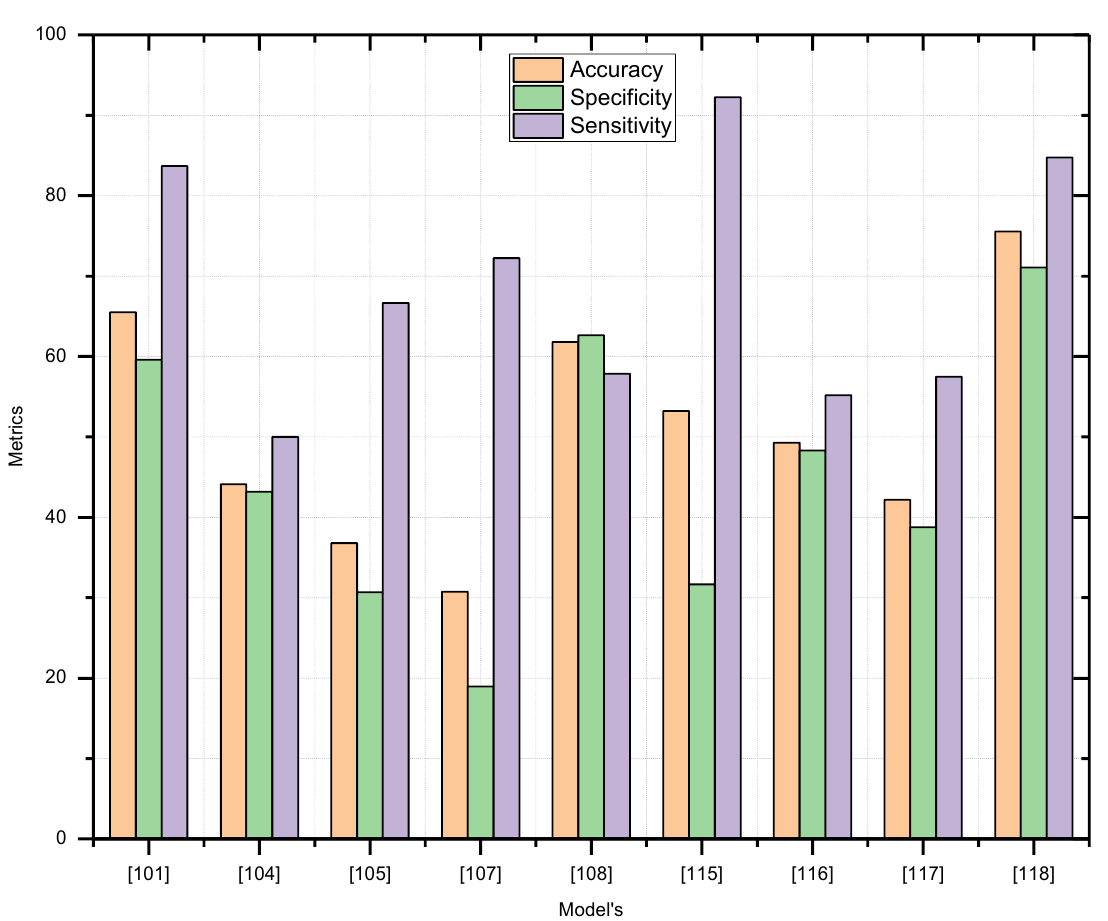}

\caption{Synopsis of \ac{CNN}-driven research in \ac{TC} diagnosis with percentages for accuracy, sensitivity, and specificity \cite{mahurkar2017normalization}, and \cite{canton2021automatic, peng2021deep, guan2019deep, teixeira2017learning, liu2019automated, qiao2021deep, zhang2020detection, tekchandani2021severity}.}

\label{fig5}
\end{figure}

\noindent \textbf{(b) CNN and RNN: } \acp{CNN}, a branch of \ac{DL} models, stand out for their remarkable capabilities in tasks such as image analysis and processing, including the classification of medical images. Their efficiency in managing data structured in grids, like images where the spatial relationship between pixels is crucial, makes them particularly suited for these tasks. The focus on \ac{CNN}-based techniques for \ac{TCD}, especially in automating nodule identification and classification in \ac{US} imagery \cite{liu2019automated}, has grown significantly. The ConvNet model, known for its reliance on convolution operations vital for image recognition tasks \cite{ha2019deep}, is a prime example of this effort. Various architectures of \ac{CNN} such as 
VGG,  AlexNet, and GoogleNet, among other \cite{himeur2023video},  are celebrated for their inclusion of convolutional, pooling, and fully connected layers. In a notable study by Li et al. \cite{li2019diagnosis}, the efficacy of \ac{CNN} models in predicting \ac{TC} was investigated, utilizing a dataset of 131,731 \ac{US} images from 17,627 patients. Xie et al. \cite{xie2019thyroid} implemented models such as Inception-Resnet, Inception, and VGG16 to differentiate malignant from benign tissues in 451 images of thyroid from the DDTI dataset, employing image augmentation to mitigate data limitations prior to classification. Moreover, Koh et al. \cite{koh2020diagnosis} assessed the diagnostic accuracy of \ac{DCNN} models against that of expert radiologists for identifying \acp{TN} in \ac{US} images, using a dataset of 15,375 images and showcasing the CNNE1 and CNNE2 models derived from \ac{DCNN} for differentiating between malignant and benign nodules. Liang et al. \cite{liang2020convolutional} introduced a \ac{DL}  based on \ac{CNN} for classifying and detecting thyroid and nodules of breast, comparing its performance with traditional \ac{US} imaging results. Figure \ref{fig5} depicts the recent advancements in classifying \ac{TCD} via \ac{CNN}-based methods.

\Acp{RNN} characterized by connections between units forming a directed graph across temporal sequences. This structure allows them to leverage internal memory, making them adept at handling inputs of varying lengths. Consequently, \acp{RNN} excel in tasks that require understanding temporal dependencies, such as speech recognition, language translation, and time-series analysis. Within the realm of thyroid carcinoma classification, \acp{RNN} offer promising capabilities for analyzing data that is sequential or time-sensitive. This includes observing the evolution of clinical symptoms over time, tracking changes in tumours using successive medical images, or studying fluctuations in gene expression associated with the onset of \ac{TC}. For example, Chen et al. (2017) utilized a hierarchical \ac{RNN} structure to categorize \acp{TN} by analyzing historical \ac{US} reports \cite{chen2017thyroid}. Their model comprises three layers of \ac{LSTM} networks trained independently. The findings from their study suggest that this hierarchical \ac{RNN} approach surpasses conventional models such as \ac{SVM} + Unigrams, \ac{SVM} + Bigrams, \ac{CNN}, and \ac{LSTM} in accuracy, computational efficiency, and robustness. These benefits are attributed to the \ac{RNN}'s memory mechanisms, which permit the retention of information from previous states through feedback loops, thereby making \Acp{RNN} highly effective for cancer detection applications.

\noindent \textbf{(c) GAN:}  \Ac{GAN} is composed of two key elements: a generator and a discriminator. The generator's function is to convert a random input vector into a data point that fits within the space of the dataset. Conversely, the discriminator serves as a binary classifier, tasked with assessing whether input data, originating either from the training dataset or produced by the generator, is genuine. \ac{GAN}s have found extensive applications in medical diagnosis, notably in the detection of \ac{TN} \cite{yoo2020generative}.

Table \ref{table:2} serves as an overview of various research initiatives aimed at identifying both benign and malignant forms of \ac{TC} It outlines the classifiers used, diseases focused on, datasets applied, research goals, and metrics for assessment. This table helps categorize the \ac{AI} techniques applied in \ac{TCD}, underlining their key roles in the domain.

\begin{table}[t!]
\leftskip=-4cm 
\caption{Summary of studies on identifying benign and malignant \ac{TC}, sorted by reference. }
\label{table:2}
\scriptsize
\begin{tabular}{
m{0.6cm}
m{0.6cm}
m{0.6cm}
m{0.6cm}
m{0.6cm}
m{0.6cm}
m{0.6cm}
m{1.5cm}
m{1.5cm}
m{1.5cm}
m{1.5cm}
m{2.5cm}
m{2.5cm}
}
\hline 

Ref. & \multicolumn{2}{c}{AI Tech.} & \multicolumn{3}{c}{Classifier} & & Objective & DD & Dataset & APP & SV \\
\cline{2-3}
\cline{4-7}
 & ML& DL & CNN & SVM & ELM & Other & & & & &   \\  \hline

\cite{sajeev2020thyroid} & & \cmark & \cmark & & & & C & TC & PD & Omics & 10068 images \\

\cite{soulaymani2018epidemiological} & & \cmark & & & & \cmark & C & TC & PD & NA & NA \\

\cite{bhalla2020expression} &  \cmark & & & \cmark & & & C & PTC & TCGA & Omics & 500 patients  \\

\cite{yadav2020prediction} &  \cmark & & & & & \cmark & C & TC & UCI & US & 3739 patients \\

\cite{zhao2015logistic} & \cmark & & & & & \cmark & C & TC & PD & US & 63 patients \\

\cite{yazdani2018factors} & \cmark & & & & & \cmark & C & TN & PD & US & 33,530 patients \\ 

\cite{rao2019thyroid} & \cmark & & & & & \cmark & C & TD & PD & US & 7200 samples  \\ 

\cite{hosseinzadeh2020multiple} & \cmark & & & & & \cmark & C & TD & UCI & US & 7200  patients \\  

\cite{chen2020diagnosis} & & \cmark & & & &\cmark & C & TN & PD & US & 1480 patients \\ 

\cite{mahurkar2017normalization} &  \cmark & & & & & \cmark & C & TC & UCI & US & 215  instances \\ 

\cite{yang2019information} &  \cmark & & & & & \cmark & C & TC & PD & US & 734 cases \\

\cite{ferreira2018autoencoders} &  & \cmark &  & & & \cmark & C & PTC & TCGA & US & 18985 features \\ 

\cite{teixeira2017learning} &  &\cmark & & & & \cmark & C & PTC & TCGA & Omics & 510 samples \\

\cite{li2019diagnosis} &  & \cmark & \cmark & & & & C & TC & PD & US & 17627 patients \\ 

\cite{xie2019thyroid} &  & \cmark & \cmark & & & & C & TC & PD & US & 1110 images \\ 

\cite{liang2020convolutional} &  & \cmark & \cmark & & & & C, P & TN & PD & US & 537 images \\ 

\cite{chen2017thyroid} &  & \cmark & & & & \cmark & C & TN & PD & US & 13592 patients  \\ 

\cite{chandel2016comparative} &  \cmark & & & & & \cmark & C & TD & PD & US & 7200 instances \\ 

\cite{ma2018efficient} & \cmark & & & & \cmark & & C & TD & UCI & US & 215 patients  \\ 

\cite{xia2017ultrasound} & \cmark & & & & \cmark & & C & TD & PD & US & 187 patients \\

\cite{dharmarajan2020thyroid} &  \cmark & & & & & \cmark & C & TC & NA & US & NA \\ 

\cite{yadav2019decision} &  \cmark & & & & & \cmark & C & TC & UCI & US & 499 patients \\ 

\cite{thomas2020aibx} & & \cmark & \cmark & & & & C & TC & PD & Omics & 482 images \\

\cite{kezlarian2020artificial} & & \cmark & \cmark & & & & NA & PTC, FTC & NA & FNAB & NA \\ 

\cite{sanyal2018artificial} &  & \cmark & \cmark & & & & C & PTC & PD & FNAB & 370 MPG \\ 

\cite{yoon2020artificial} &  & \cmark & \cmark & & & & P & PTC & PD & FNAB & 469 patients \\ 

\cite{nguyen2020ultrasound} & & \cmark & \cmark & & & & C & TC & DDTI & US & 298 patients \\ 

\cite{liu2017classification} &  & \cmark & \cmark & & & & C & TC & PD & US & 1037 images \\ 

\cite{abdolali2020automated} &  & \cmark & \cmark & & & & P & TN & PD & US & 80 patients \\ 

\cite{li2018fully} &  & \cmark & \cmark & & & & P & TN & PD & US & 300 images \\ 

\cite{kim2016deep} &  & \cmark & \cmark & & & & C & TC & PD & US & 459 labeled \\ 

\cite{ma2019thyroid} & & \cmark & \cmark & & & & C & TD & ImageNet & US & 2888 samples  \\ 

\cite{chai2020artificial} &  & \cmark & & & & \cmark & NA & TC & NA & US & NA \\ 

\cite{song2019ultrasound} &  & \cmark & & & & \cmark & C & TC & PD & US & 1358 images \\ 

\cite{barczynski2020clinical} & \cmark & & & & & \cmark & C & TC & PD & Surgery & 50 patients \\ 

\cite{choi2017computer} & \cmark & & & & & \cmark & C & TC & PD & US & 89 patients \\ 

\cite{fragopoulos2020radial} & \cmark & & & & & \cmark & C & TD & PD & Cyt & 447 patients \\ 

\cite{savala2018artificial} & \cmark & \cmark & & & & \cmark & C & FTC & PD & FNAB & 57 smears \\ 

\cite{li2021artificial} &\cmark & \cmark&  & & & \cmark & NA & FTC & NA & FNAB & NA \\ 

\cite{zhao2019assessment} &\cmark & \cmark & & & & \cmark & P & TC & TCGA & Hist& 482 samples \\ 

\cite{wildman2019using} & \cmark & \cmark & & & & \cmark & C & TC & PD & FNAB & 1264 patients \\

\cite{wang2019automatic} &\cmark & \cmark  & & & & \cmark & C & TN & PD & US & 276 patients \\ 

\cite{ozolek2014accurate} &  \cmark & & & & & \cmark & C, P & FTC & PD & Hist & 94 patients \\ 

\cite{zhu2019deep} &  \cmark & & & \cmark & & & C & TN & PD & US & 467 TN  \\

\cite{dolezal2020deep} &  & \cmark & & & & \cmark & PTC & TCGA & Omics & 115 slides & NA \\ 

\cite{daniels2020machine} &  \cmark & & & & & \cmark & C & TN & PD & Omics & 121 patients \\ 

\hline
\end{tabular}
\begin{flushleft}
Abbreviation: Application (APP), Detected disease (DD), Subjects for validation (SV), Private data (PD), Classification (C), Prediction (P), Segmentation (S), Cytopathological (Cyt), Histopathological (Hist),
Microphotographs (MPG).
\end{flushleft}
\end{table}

\subsection{Comparative analysis and discussion} 

This section commits to an exhaustive examination of \ac{AI} models' proficiency in identifying thyroid carcinoma. We aim to scrutinize not merely their statistical precision but also their efficacy in practical clinical environments, and their contribution to the overarching clinical decision-making framework. Furthermore, this exploration delves into the potential biases embedded in \ac{AI} models, seeking to unveil how they might inadvertently amplify existing healthcare inequities. By contrasting \ac{AI}-enabled methodologies with conventional diagnostic tactics, we aspire to glean deeper insights into their relative effectiveness.

\begin{figure}[ht!]
\centering
\includegraphics[scale=0.7]{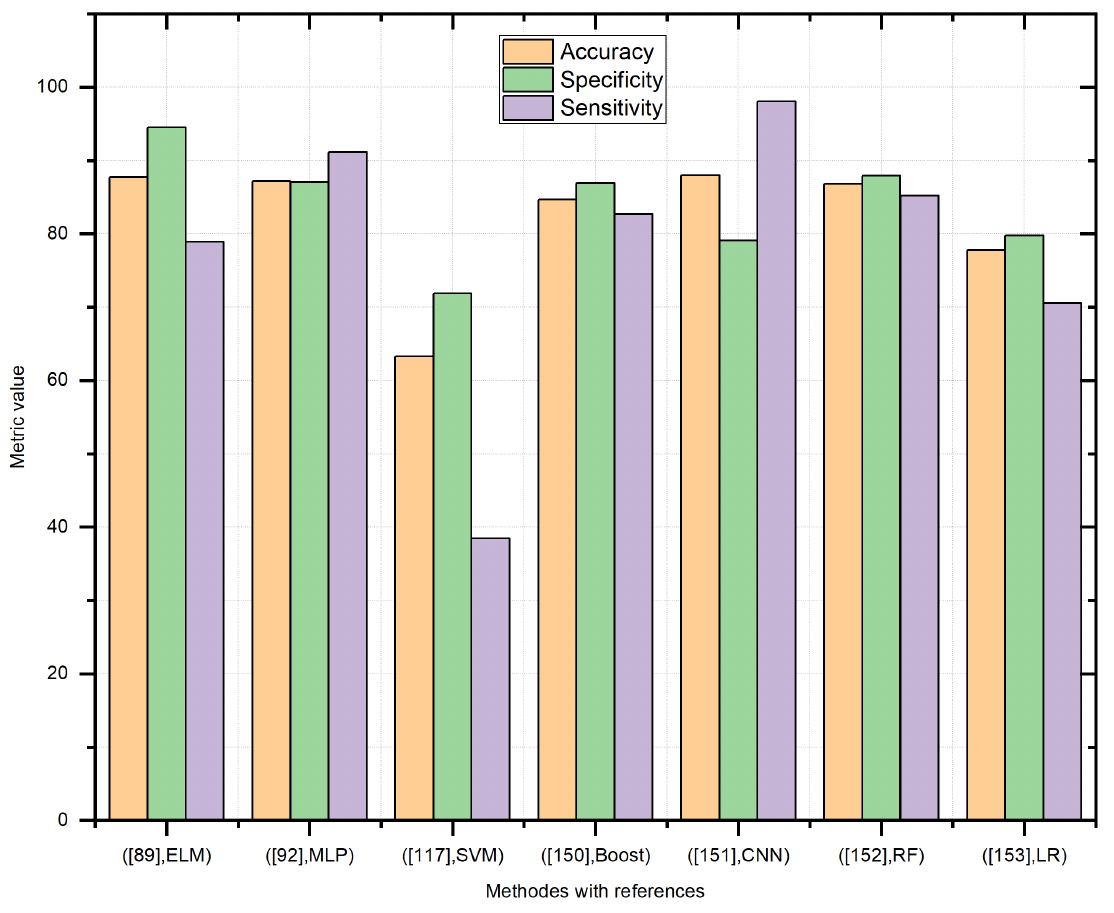}
\caption{Performance assessment of \ac{TC} frameworks in percentages (\%) for private \ac{TCD} \cite{mehta2019high}, \cite{guo2019xgboost}, \cite{zhang2020detection}, \cite{tran2023video}, \cite{gu2019prediction, colakoglu2019diagnostic, park2021combining}.}
\label{fig15}
\end{figure}

\begin{figure}[t!]
\centering
\includegraphics[scale=0.7]{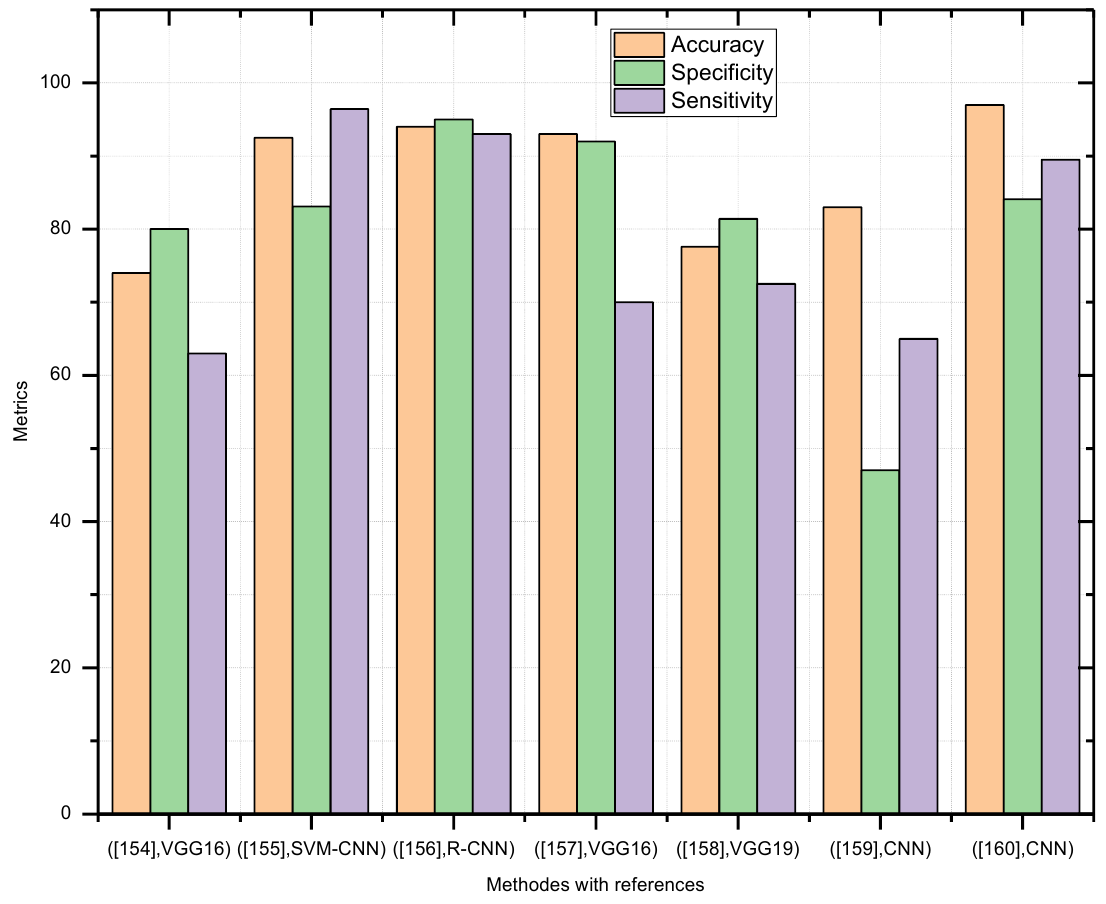}
\caption{Performance assessment of TC frameworks in percentages (\%) for \ac{US} \ac{TCD} \cite{mughal2018novel}, \cite{vaswani2017attention}, \cite{lu2020application, kwon2020ultrasonographic, chan2021using, wang2021radiomic, zhou2020online}.}
\label{fig16}
\end{figure}

The reported metrics of \ac{AI} models, such as accuracy, sensitivity, and specificity, can vary significantly across academic publications due to factors like the choice of dataset, data quality, and the methodological approach utilized. The performance of \ac{AI} models in controlled experimental setups may not accurately represent their effectiveness in actual clinical scenarios. Variables including data discrepancies, lack of complete data, and evolving clinical conditions can substantially influence outcomes. Therefore, assessing a model's flexibility and reliability in diverse conditions is crucial. Table \ref{tab8} provides an overview of the performance indicators for various \ac{AI}-enhanced \ac{TCD} frameworks, presented in percentage (\%) terms, across multiple models and data sources. Additionally, Figures \ref{fig15} and \ref{fig16} display these performance indicators, also in percentage (\%) terms, with a focus on private datasets and \ac{US} imaging data, respectively.

\begin{table}[ht!]
\leftskip=-4cm 
\caption{Assessment of the effectiveness of different
\ac{TCD} classification schemes, measured in percentages (\%).}
\label{tab8}
\scriptsize
\begin{tabular}{
m{6mm}
m{35mm}
m{35mm}
m{15mm}
m{15mm}
m{15mm}
m{15mm}
}
\hline
Ref. &  AI Model & Dataset & SEN & SPE & ACC & AUC\\
\hline
\cite{zhang2019machine} & \ac{RF} & US  & - & - & - & 94.00 \\

\cite{peng2021deep} & ThyNet & PD & - & - & - & 92.10 \\
\cite{li2019diagnosis} & DCNN &SGI & 93.00 & 86.00 & 89.00 & - \\
\cite{guan2019deep} & VGG-16 & CI & - & - & 97.66 & - \\
\cite{zhu2019deep} & \ac{DNN} & ACR T & - & - & 87.20 & - \\
\cite{gu2019prediction} & CNN & CT image  & 93.00 & 73.00 & 84.00 & - \\
\cite{ma2017ultrasound} & CNN & DICOM & - & 91.50 & - & - \\
\cite{chi2017thyroid} & Fine-Tuning DCNN & PD & - & - & 99.10 & - \\
\cite{duc2022ensemble} & Ensemble \ac{DL} & CI & - & - & 99.71 & - \\
\cite{ouyang2019comparison} & k-SVM & US & - & - & - & 95.00\\
\cite{zhao2021comparative} & \ac{SVM} \ac{RF} & US & - & - & - & 95.10 \\
\cite{vadhiraj2021ultrasound} & \ac{ANN} \ac{SVM} & US & - & 96.00 & - & - \\
\cite{gild2022risk} & \ac{RF} & US & - & - & - & 75.00 \\
\cite{ma2017pre} & CNN & DICOM & 82.40 & 85.00 & 83.00 & - \\
\cite{zhu2017image} & ResNet18-based &  PD& - & - & 93.80 & - \\
\cite{gao2018computer} & multiple-scale CNN & PD & - & - & 82.20 & - \\
\cite{zuo2018extraction} & Alexnet CNN & PD & - & - & 86.00 & - \\
\cite{zhu2021efficient} & CNN (BETNET) & US & - & 98.30 & - & - \\
\cite{kim2022deep} & ResNet & T & - & 75.00 & - & - \\
\cite{lee2019application} & Xception & CT images & 86.00 & 92.00 & 89.00 & - \\
\cite{tsou2019mapping} & Google inception v3 & HPI & - & - & 95.00 & - \\
\cite{kim2021convolutional} & CNN & T  & 81.80 & 86.10 & 85.10 & - \\
\cite{wu2021deep} & CNN & T  & 78.00 & 85.00 & 82.10 & - \\
\cite{jin2020ultrasound} & CNN & T  & 80.60 & 80.10 & 80.30 & - \\
\cite{park2019association} & CNN & US  & - & - & 77.00 & - \\
\hline
\end{tabular}
\begin{flushleft}
 Abbreviations:  TIRADS (T), Cytological images (CI), Sono graphic images (SGI), Histo pathology images (HPI)   
\end{flushleft}
\end{table}

\subsection{Case study example}

To exemplify the approaches adopted in the literature for \ac{TCD} and the utilization of \ac{AI} in classifying cancer types, we provide a simplified example. The pattern recognition process entails training a neural network to accurately classify input patterns into specific target classes. Following training, the network becomes capable of categorizing model. In this part, we demonstrate an example of categorizing \ac{TC} into benign, malignant, or normal based on a collection of characteristics according to the \ac{TIRADS}.

The dataset obtained from the \ac{UCI} \ac{ML} repository has been used \cite{murphy1994uci}, which categorizes patients visiting a clinic into three distinct groups: normal, hyperfunctioning, or subnormally functioning.   The dataset is structured into \ac{TT} and \ac{TI} as detailed below: (i) \ac{TI} consists of a 21$\times$7,200 matrix that describes 7,200 patients through 15 binary and 6 continuous attributes. (ii) \ac{TT} is a 3$\times$7,200 matrix with class vectors, distributing each patient input into one of the three categories: (1) Hyperfunctioning, (2) Normal, not suffering from hyperthyroidism, and (3) Subnormal functioning.

In this neural network setup, the dataset is divided into 5,040 samples for training, 1,080 samples for validation, and 1,080 samples for testing purposes. The network undergoes training to minimize the error between the thyroid inputs and targets until it achieves the desired target objective. If the \ac{ER} fails to decrease and training progress stalls, training with the training data is stopped, and the validation data is utilized for additional evaluation. Subsequently, the testing data is employed to assess the accuracy of the trained model.

\begin{figure}[t!]
\centering
\includegraphics[width=1\columnwidth]{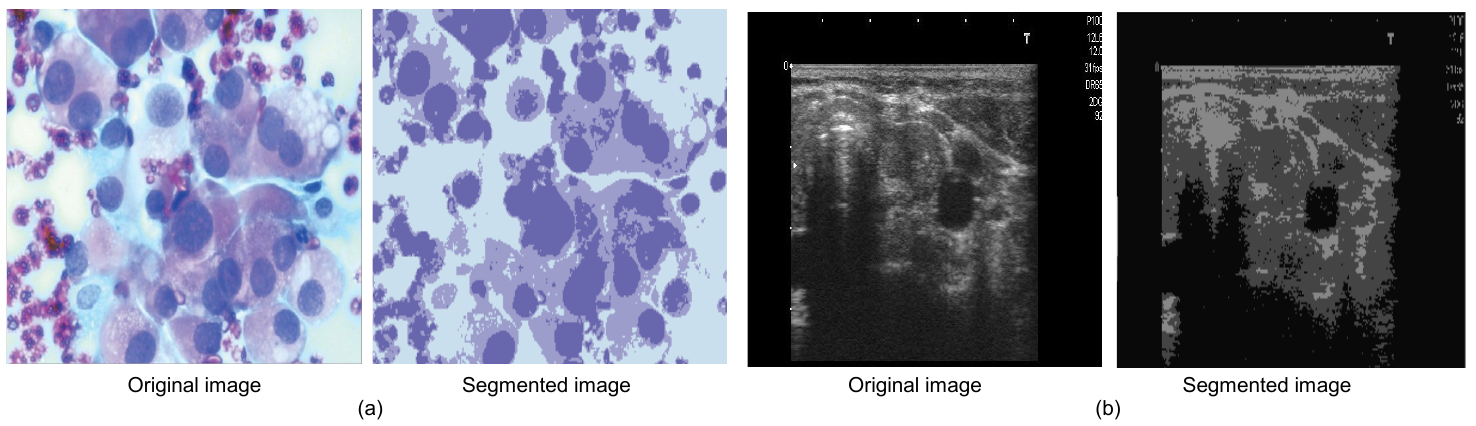}
\caption{ Thyroid segmentation example employing the K-Means method. (a) Imaging of medullary \ac{TC}; (b) \ac{US} imaging of the thyroid.}
\label{fig11}
\end{figure}

\begin{figure}[t!]
\centering
\includegraphics[width=0.6\columnwidth]{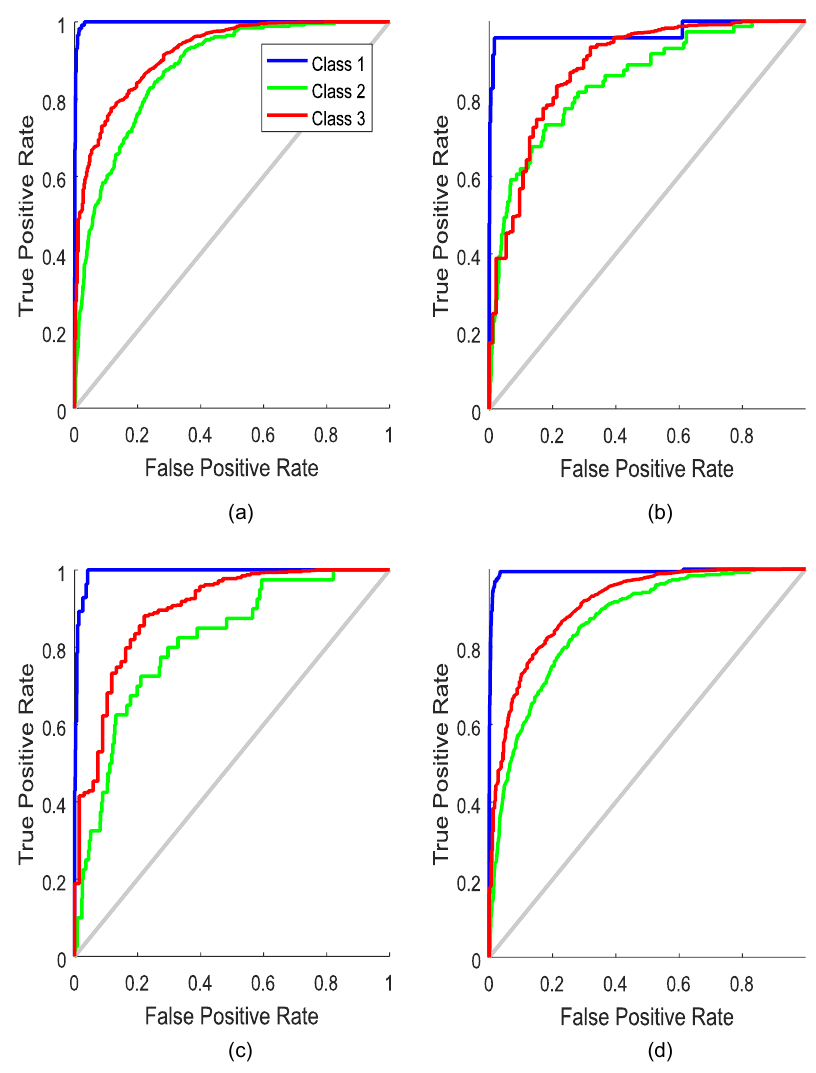}
\caption{An example of \ac{TCD} classification using \ac{ROC} metric. (a) Training \ac{ROC}; (b) Validation \ac{ROC}; (c) Test \ac{ROC}; (d) All ROC}
\label{fig10}
\end{figure}

Figure \ref{fig11} illustrates an instance of thyroid segmentation in \ac{US} images utilizing k-means clustering, with three clusters selected for demonstration. K-means clustering is widely employed for such purposes. In this illustration, a network featuring 10 hidden layer neurons has been employed, 21 input features, and 3 output classes. Following the model simulation, the percentage error is computed, yielding values of 5.337\% for training, 7.407\% for validation, and 5.092\% for testing. The overall recognition rate stands at 94.4\%, with an overall error rate of 5.6\%. The \ac{ROC}  curve is presented in Figure \ref{fig10}. This example showcases the application of \ac{AI}  in \ac{TCD}  classification, achieving a high recognition rate with the provided dataset.

\section{Advanced TCD using ViT and LLM }
\label{sec4}

\Ac{ViT} and their advanced version, \acp{LLM}, have emerged as cutting-edge techniques in various AI-based biomedical tasks. Recently, researchers have begun applying these models to \ac{TCD}. The following subsections provide a review of recent studies, along with a performance summary presented in Table \ref{table:7}.

\subsection{ViT-based TCD methods}
\color{black}

\begin{figure}[t!]
\centering
\includegraphics[width=1\columnwidth]{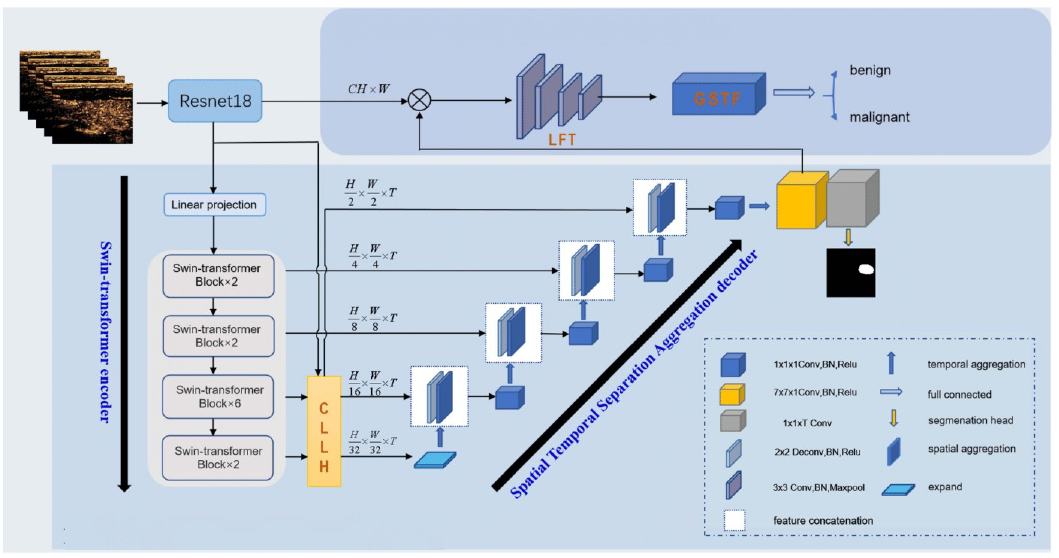}\\
(a)\\
\includegraphics[width=10cm, height=3.5cm]{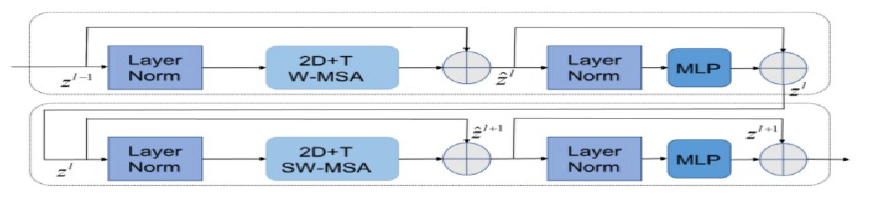}\\
(b)\\
\caption{ (a) The general framework of the suggested HEAT-Net, (b) Details of the Swin-Transformer block  \cite{chen2023joint}}.
\label{fig7}
\end{figure}

\Acf{ViT} models are a type of \ac{DL} model introduced in \cite{vaswani2017attention}. They have since become the foundation for many state-of-the-art \ac{NLP} and \ac{ML} models. Transformers are designed to handle sequential data, making them suitable for various tasks beyond \ac{NLP} as well \cite{kheddar2024transformers,djeffal2023automatic,kheddar2024automatic}. Transformers in \ac{TCD} enable efficient analysis of medical data, including literature and patient records, offering high accuracy. They uncover patterns, risk factors, and diagnostic clues, aiding early detection. This technology enhances diagnosis speed, fuels data-driven research, and promises improved patient care and oncology advancements \cite{tran2023video, chi2023hybrid}. Several studies have proposed the use of \ac{DL} models, particularly Transformer-based models, for detecting \ac{TC} from \ac{US} images. For example, researchers in \cite{sharma2023framework} developed a diagnostic system using \ac{DL} (Deit, Swin Transformer, and Mixer-MLP) and metaheuristics to improve thyroid abnormality detection.  The method in \cite{sharma2023framework} ranked the models, leading to the selection of the best-performing models for ensemble learning. The optimization-based feature selection and random forest model achieved high accuracy on \ac{US} and histopathological datasets, surpassing existing methods. This innovative approach eases the burden on healthcare professionals by enhancing \ac{TC} diagnosis. The study \cite{pathak2023extracting}, addressed the challenge of extracting important \ac{TN} characteristics from clinical narratives in \ac{US} reports using \ac{NLP}. A team of experts developed annotation guidelines and tested five Transformer-based \ac{NLP} models. Their GatorTron model, trained on a substantial text corpus, outperformed others, achieving the best F1-scores for extracting 16 \ac{TN} characteristics and linking them to nodules. This pioneering work enables improved documentation quality of thyroid \ac{US} reports and enhances patient outcomes assessment through electronic health records analysis. In \cite{bi2023bpat}, the study introduces a novel \ac{BPAT-UNet} for precise \ac{US} \ac{TN} segmentation. This network incorporates a \ac{BPSM} for boundary refinement and an \ac{AMFFM} for handling various scales of features. Additionally, an \ac{ATM} improves boundary constraints and small object detection. Results demonstrated significantly improved segmentation accuracy compared to classical networks, achieving Dice similarity coefficients of 85.63\% and 81.64\% and HD95 values of 14.53 and 14.06 on private and public datasets, respectively. Chen et al. \cite{chen2023joint}, introduce Trans-CEUS, a spatial-temporal Transformer-based model for real-time microvascular perfusion analysis using \ac{CEUS} as it shown in Figure \ref{fig7}. It combines dynamic Swin-Transformer and collaborative learning to accurately segment lesions with unclear boundaries, achieving a Dice similarity coefficient of 82.41\%. The model also attains a high diagnostic accuracy of 86.59\% for distinguishing malignant and benign \ac{TN}. This pioneering research highlights the effectiveness of Transformers in \ac{CEUS} analysis and offers promising outcomes for \ac{TN} segmentation and diagnosis from dynamic \ac{CEUS} datasets.

\ac{DL} has been instrumental in medical image segmentation, particularly for thyroid glands in \ac{US} images. However, existing models face issues like the loss of low-level boundary features and limitations in capturing contextual features. In response, a hybrid Transformer UNet is introduced in \cite{chi2023hybrid}. It combines a 2D Transformer UNet with a multi-scale cross-attention Transformer and a 3D Transformer UNet with self-attention to improve representation and contextual information. The end-to-end network was evaluated on thyroid segmentation datasets, outperforming other methods in benchmark tests. The method shows promise for thyroid gland segmentation in \ac{US} sequences. Dataset classification involves predicting a single label from sets with multiple instances, like pathology slides or medical text data. State-of-the-art methods often use complex attention architectures to model set interactions. However, when labeled sets are limited, as in medical applications, these architectures are challenging to train. To tackle this issue, a kernel-based framework is introduced in \cite{dov2021affinitention}, connecting affinity kernels and attention architectures. This leads to simplified "affinitention" nets, which are applied to tasks like Set-Cifar10 classification, thyroid malignancy prediction, and patient text triage. Affinitention nets deliver competitive results, outperforming heuristic attention architectures and other methods. 
Jerbi et al. in 2023 \cite{jerbi2023automatic}, incorporating \acp{CNN} and \ac{ViT}, was employed to classify thyroid \ac{US} images as either malignant or benign. A deep convolutional \ac{GAN} was used to address data scarcity and imbalance. Various models, including VGG16, EfficientNetB0, ResNet50, \ac{ViT}-B16, and hybrid \ac{ViT}, were trained with both softmax and \ac{SVM} classifiers. The hybrid \ac{ViT} model, with \ac{SVM} classification, outperformed others, achieving a 97.63\% accuracy, showing promise for aiding doctors in diagnosing thyroid patients more effectively.

\begin{figure}[t!]
\centering
\includegraphics[width=1\columnwidth]{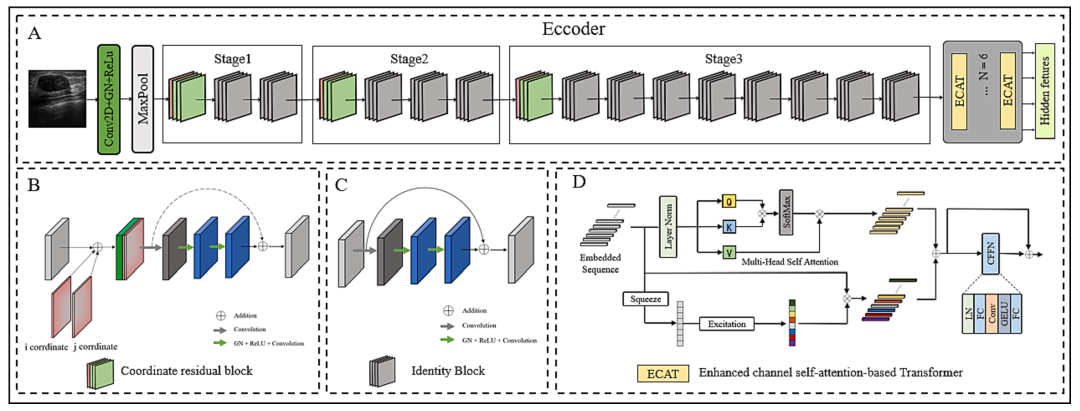}
\caption{An example of the basic structure of encoder HEAT-Net for \ac{TC} segmentation \cite{jiang2023hybrid}.}
\label{fig8}
\end{figure}

\begin{figure}[t!]
\centering
\includegraphics[width=0.8\columnwidth]{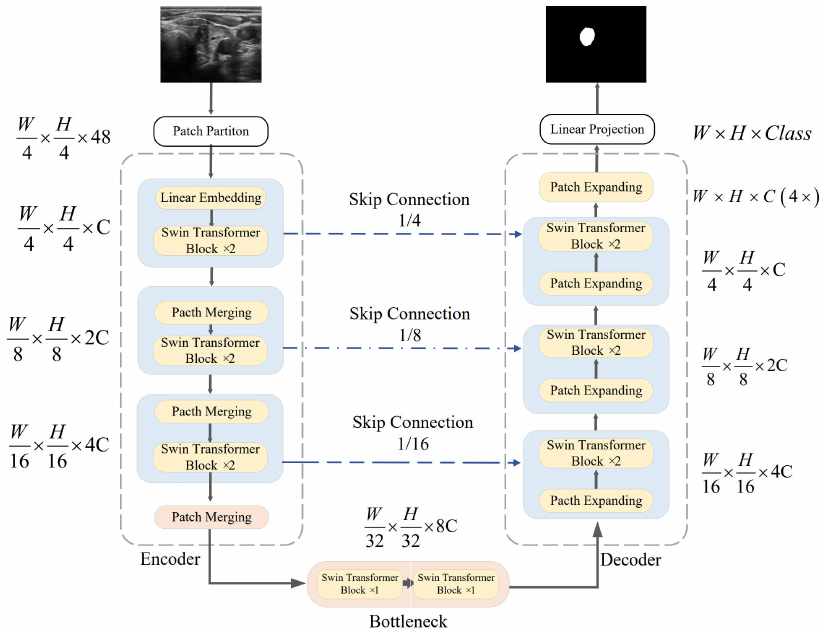}
\caption{ The Swin-Unet architecture \cite{ma2023amseg}.}
\label{fig9}
\end{figure}

In \cite{jiang2023hybrid}, the authors introduce a novel U-shape segmentation model (Figure \ref{fig8}) combining CNN and Transformer structures to integrate local and long-range information. It uses coordinate residual blocks (CdRB) to encode position data, channel-enhanced self-attention-based Transformers for global feature enhancement, and a dual attention module for feature correlation and edge accuracy. The method outperforms state-of-the-art methods across various datasets, demonstrating adaptability and robustness in \ac{US} image segmentation, potentially serving as a general segmentation tool. The work \cite{li2023fusing} addresses the challenge of accurately diagnosing malignant \ac{TN} through \ac{US} imaging. Existing \ac{CAD} methods often struggle to maintain precise shape information and capture long-range dependencies. The proposed Transformer fusing CNN Network utilizes a large kernel module in a CNN branch for shape feature extraction and an enhanced Transformer module in another branch for remote pixel connections. A Multiscale fusion module integrates feature maps from both branches. Comparisons with other methods demonstrate the superiority of the proposed scheme and its effectiveness in nodule segmentation. Hypoparathyroidism is a major concern post-TC surgery, affecting patients' quality of life. Identifying and locating parathyroid glands via \ac{US} images before surgery can help protect them. In \cite{liu2023dca} a dual-branch contextual-aware network with Transformer is proposed to reduce hypoparathyroidism incidence. It combines a Transformer for global context extraction and a feature encoder branch for local feature aggregation. A channel and spatial fusion module integrates information from both branches. The proposed method effectively addresses detail loss, establishing global and local feature dependencies. Experiments with an \ac{US} image dataset demonstrate superior performance compared to existing methods. The thyroid gland plays a crucial role in regulating the human body's functions, making the identification of \ac{TN} from \ac{US} images important for medical diagnosis. However, the automatic segmentation of these nodules is challenging due to their heterogeneous appearance and background similarities. This framework  \cite{ma2023amseg} presents a novel framework AMSeg based on Swin-Unet architecture presented in Figure \ref{fig9}, which employs multi-scale anatomical features and late-stage fusion through adversarial training to address these challenges. Experimental results demonstrate the superiority of AMSeg in \ac{TN} segmentation, achieving high dice, Hd95, Jaccard, and precision values. This end-to-end network offers promise for clinical applications, potentially replacing manual segmentation methods.

In \cite{nam2023bert}, the authors harnessed \ac{NLP} with a \ac{BERT} classifier to analyze unstructured clinical text data pertaining to recurrent papillary \ac{TC} diagnosis. The \ac{BERT} model achieved exceptional performance, boasting a 98.8\% accuracy in binary recurrence classification. This approach streamlines the handling of unstructured patient information, eliminating the need for labour-intensive data refinement, and holds significant promise for training \ac{AI} models in healthcare. The variability in features between benign and malignant \ac{TN}, particularly in \ac{TIRADS} level 3, can lead to inconsistent diagnoses and unnecessary biopsies. To address this in \cite{sun2023classification}, \ac{ViT}-based TCD, utilizes contrast learning to enhance diagnostic accuracy and biopsy specificity. By incorporating global and local features, this model minimizes the distinction between nodules of the same category. Test results indicate an accuracy of 86.9\%, outperforming classical \ac{DL} models. It offers automatic classification of \ac{TIRADS} 3 and malignant nodules in \ac{US} images, promising improved \ac{CAD} and precise analysis. \Ac{OCT} can aid in distinguishing normal and diseased thyroid tissue during surgery, but interpreting this type of image is challenging. Similarly, \cite{tampu2023diseased} explored various \ac{DL} models for classifying thyroid diseases using 2D and 3D \ac{OCT} data from 22 surgical patients with thyroid pathologies. The 3D \acp{ViT} model achieved the best performance, with an accuracy of 0.90 for normal versus abnormal classification. Custom models also excelled on open-access datasets. These findings suggest that combining \ac{OCT} with \ac{DL} can enable real-time, automatic identification of diseased tissue during thyroid surgery. 

Accurate segmentation of \acp{TN} in \ac{US} images is crucial for early \ac{TC} diagnosis. Addressing the challenges posed by weak image edges and complex thyroid tissue structure, the study \cite{tao2022local} introduces the LCA-Net method. It combines local features from \acp{CNN} and global context from Transformers, improving edge information capture. The model incorporates specific modules to handle different nodule sizes and positions, enhancing generalization. LCA-Net outperforms existing models on public datasets, demonstrating its potential for precise \ac{TN} diagnosis in clinical settings.  In this study \cite{wang2023shared}, the authors focus on improving the prediction of lymph node metastasis in papillary thyroid carcinoma by combining \acp{WSI} and clinical data. They introduce a Transformer-guided multi-modal multi-instance learning framework that effectively groups high-dimensional \acp{WSI} into low-dimensional feature embeddings and explores shared and specific features between modalities. The approach achieved an impressive \ac{AUC} of 97.34\% on their dataset, outperforming state-of-the-art methods by 1.27\%, highlighting its potential in improving precision medicine decisions based on multi-modal medical data fusion. Diagnosing lymph node metastasis in papillary thyroid carcinoma typically relies on analyzing large \acp{WSI}. To enhance accuracy, a novel Transformer-guided framework is introduced in \cite{wang2022lymph}, leveraging Transformers in three critical aspects. It incorporates a lightweight feature extractor, a clustering-based instance selection scheme, and a Transformer-MIL module for effective feature aggregation. The model further benefits from an attention-based mutual knowledge distillation paradigm. Experimental results on a \ac{WSI} dataset outperform state-of-the-art methods by a significant margin, achieving a 2.72\% higher \ac{AUC}. Xiao et al. in \cite{xiao2022contrast} aims to address the challenges of diagnosing \ac{TC}, particularly in cases where \ac{US} images suffer from noise and artefacts, leading to a certain misdiagnosis rate in clinical practice. They highlight the need for further diagnosis using plain and contrast-enhanced \ac{CT} scans. While plain \ac{CT} provides valuable information, contrast-enhanced \ac{CT} offers better contrast and can reflect organ margin erosion, a crucial symptom for \ac{TC} diagnosis. However, the latter relies on the use of a contrast agent and exposes the patient to ionizing radiation. To mitigate these challenges, the authors propose an improved Unet architecture. Their approach involves using a convolutional Transformer module to learn global information from high-dimensional features. They also incorporate a texture feature module to extract local texture information from plain \ac{CT} scans and integrate edge information obtained from superpixels as prior knowledge. The ultimate goal is to generate enhanced \ac{CT} images with clear texture and higher quality, providing a valuable tool for \ac{TC} diagnosis without the need for contrast agents and ionizing radiation. Histopathological images carry valuable information for tumour classification and disease prediction, but their size hinders direct use in \acp{CNN}. This study \cite{yin2022pyramid} introduces Pyramid Tokens-to-Token \ac{ViT}, a lightweight architecture with multiple instance learning based on the \acp{ViT}. The method uses the Tokenization technique for feature extraction to reduce model parameters. It also incorporates an image pyramid to capture local and global features, significantly reducing computation. Experiments on thyroid pathology images yield superior results compared to CNN-based methods, balancing accuracy and efficiency. The authors in \cite{yu2023pyramid} aim to utilize multi-instance learning for the diagnosis of \ac{TC} based on cytological smears. These smears lack multidimensional histological features, necessitating the mining of contextual information and diverse features for improved classification performance. To address these challenges, they introduce a novel algorithm called PyMLViT, which consists of two core modules. First, the pyramid token extraction module is designed to capture potential contextual information from smears. This module extracts multi-scale local features using a pyramid token structure and obtains global information through a \acp{ViT} structure with a self-attention mechanism. Second, they construct a multi-loss fusion module based on the conventional multi-instance learning framework. To enhance the diversity of supervised information, they carefully allocate bag and patch weights and incorporate slide-level annotations as pseudo-labels for patches during training. In Table \ref{table:7}, Transformer-based models' performance (in \%) for \ac{TC} diagnosis is summarized.

\begin{table}[tbph]
\leftskip=-4cm 
\scriptsize
\caption{Summary of Transformer and LLM-based models' performance (in \%) for \ac{TC} diagnosis.}
\label{table:7}

\begin{tabular}{
m{0.4cm}
m{2cm}
m{0.5cm}
m{0.5cm}
m{0.5cm}
m{0.5cm}
m{0.5cm}
m{0.5cm}
m{8.5cm}
}
\hline
Ref. & Transformer & Task & ACC & SEN & SPE & AUC & F1-score & Improvement\\
\hline
\cite{sharma2023framework} & 
  Swing &
C  & 
99.13 &
-- &
-- &
99.13 &
98.82 & 
The diagnosis accuracy and the extract features of thyroid abnormality detection has been improved. The redundant features has reduced to avoid the overfitting. \\ 
\cite{pathak2023extracting} & 
BERT, RoBERTa, LongFormer, DeBERTa, and GatorTron &
C & 
97.10&
98.80&
92.80&
--&
96.50&
The proposed model can achieve satisfactory classification accuracy and identify a large number of characteristics comparable to experienced radiologists and can save time and effort as well as deliver potential clinical application value \\ 

\cite{chen2023joint} & 
Trans-CEUS &
S& 
86.59&
--&
--&
--&
--&
Demonstrated significant improvement when compared to previous approaches, showcasing its effectiveness in the tasks of lesion segmentation and \ac{TN} diagnosis.\\ 
\cite{dov2021affinitention} & 
Network&
C& 
--&
--&
--&
91.3&
--&
Attention nets outperform complex attention-based architectures and other competing methods in tasks such as thyroid malignancy prediction.\\ 
\cite{jerbi2023automatic} & 
\acp{ViT} &
C&
97.63&
--&
--&
--&
96.67&
The \ac{SVM} classification produces better performance than the Softmax classification for all of the models with the Hybrid \ac{ViT}\\ 

\cite{nam2023bert} & 
BERT &
C& 
--&
--&
--&
--&
88.00&
Analyze unstructured clinical text information on the diagnosis of the recurrent PTC efficiently.\\ 
\cite{sun2023classification} & 
\ac{ViT} &
C& 
86.90&
87.10&
86.10&
--&
92.4&
Improve accuracy of diagnosis and specificity of biopsy recommendations. Minimize the representation distance between nodules of the same category\\ 
\cite{tampu2023diseased} & 
3D vision  &
C & 
90.00&
--&
--&
--&
--&
Efficiently classifying thyroid diseases \\ 
\cite{wang2023shared} & 
GMMMIL &
P & 
93.88&
--&
--&
97.34&
94.65&
Experimental results on the collected lymph node metastasis dataset demonstrate the efficiency of the proposed method\\ 
\cite{wang2022lymph} & 
Tiny-ViT &
P& 
--&
--&
--&
98.35&
92.97&
Improve predict lymph node metastasis from \acp{WSI} efficiently using a novel Transformer-guided.\\ 

\cite{yin2022pyramid} & 
T2T-ViT &
C & 
86.60&
--&
--&
--&
--&
The model parameters are reduced , and the model performance and computation are greatly improved compared with CNN. \\ 

\cite{yu2023pyramid} & 
PyMLViT & C &  87.50& --& --& --& --& Optimize the training process of the network.\\

\cite{lee2023development} &  FastChat-T5&  MQA&  88.86&- &- &- &- & Significant reduction in time for data extraction.\\
\cite{raghunathan2025can} & GPT-4 & MQA & 95.25  &- &- &- &- & GPT-4 responses scored highest for accuracy, quality, and empathy compared to GPT-3.5 and doctors. \\
\cite{wang2024assessing} &  GPT-4 & D & 90.00 &- & -&- &- & The diagnosis closely resembles human reports. \\
\cite{wu2024collaborative} & ChatGPT-4 &D & 86.00  &- &- &- &- & Optimal performance in \ac{US} diagnosis of thyroid nodules.\\
\cite{shah2024endogpt} & GPT-4o  & D  & 93.00  &- &- &- &- & High accuracy in diagnosis and management of thyroid nodules. \\
\cite{yao2024ai} &  LlaMA2-13B & D & 87.50 &  86.20 & 88.30& 90.90 & 85.00& Diagnosis accuracy surpassed standalone \ac{CAD} models and human performance. \\
\cite{zhang2024berttcr} & BertTCR & P &  96.60 & 100 &100 &100 &95.80 & High performance in predicting thyroid cancer-related immune.\\

\hline 
\end{tabular}%
\begin{flushleft}
Abbreviations: segmentation (S); Classification (C);  Prediction (P); Medical question answering (MQA); Diagnosis (D).   
\end{flushleft}
\end{table}

\subsection{LLM-based TCD methods}
The \acp{LLM} are advanced \ac{NLP} model,  trained on vast datasets to process, understand, and generate human-like text. Using \ac{DL} and Transformer architectures, \ac{LLM}s excel in tasks like answering questions, summarizing, translating, and creating content.  In \ac{TCD}, \ac{LLM}s could be fine-tuned to assist in analyzing patient data, diagnostic reports, and medical literature, identifying patterns, and offering insights for early detection and personalized treatment. Their ability to process complex medical information enhances diagnostic accuracy, supports clinicians, and improves patient outcomes, making them invaluable in advancing thyroid cancer care and treatment. Several studies have suggested using \ac{LLM} models for identifying thyroid cancer from \ac{US} images. For instance, researchers in  \cite{lee2023development} evaluate a privacy-preserving \ac{LLM} for extracting critical clinical information from thyroid cancer pathology reports. Using FastChat-T5, the model answered 1,008 questions about staging and recurrence risk across 84 reports. Concordance rates between the \ac{LLM} and human reviewers averaged 89\%, with the \ac{LLM} completing tasks significantly faster (19.6 minutes vs. 206.9 minutes). While accurate for binary questions, challenges arose in complex queries. The findings highlight the potential of tailored \ac{LLM}s for efficient, privacy-compliant clinical data extraction. Moving on,  Raghunathan et al. \cite{raghunathan2025can} evaluate \ac{LLM}s, including ChatGPT-3.5 and GPT-4, in addressing thyroid disease patient queries compared to verified doctors. Using a 4-point Likert scale, the proposed \ac{LLM}s outperformed physicians in accuracy, quality, and empathy. GPT-4 scored highest across metrics. The findings highlight \ac{LLM}s' potential to enhance patient communication, reduce clinician workload, and mitigate burnout by providing accurate and empathetic answers to complex medical questions in thyroid care. Wu et al. \cite{wu2024collaborative} evaluated the feasibility of leveraging \ac{LLM}s like  ChatGPT 4.0 to enhance thyroid nodule diagnosis using standardized reporting (TI-RADS) and pathology as the reference standard. Among 1161 ultrasound images analyzed, ChatGPT 4.0 outperformed others in consistency and diagnostic accuracy, especially when combined with image-to-text strategies. It matched or exceeded human-\ac{LLM} interactions and showed potential to improve diagnostic efficiency while maintaining interpretability. Differently, Shah et al. \cite{shah2024endogpt} presented EndoGPT, a \ac{LLM}-based tool developed for thyroid nodule management using GPT-4o, prompt engineering, and knowledge retrieval. Tested on 50 patient scenarios, it achieved a high overall concordance with expert surgeon plans, excelling in diagnosis and operational decisions, though less so in operation type (69\%). While not a replacement for clinicians, EndoGPT highlights the potential of \ac{LLM}s in aiding medical decision-making, education, and enhancing accessibility to clinical guidelines. Similarly,  \cite{yao2024ai} introduced AIGC-\ac{CAD} model for thyroid nodules. Inspired by ChatGPT, it integrates human-computer interaction to enhance diagnostic accuracy using 19,165 ultrasound cases. By combining \ac{DL} models and semantic understanding, the model provides transparent diagnostic rationales and improves physician confidence. The model enhances junior radiologists' sensitivity and specificity by over 20\%, bridging skill gaps. Its explainable and interactive features mark a paradigm shift in \ac{CAD} applications. Wang et al. \cite{wang2024assessing} evaluates GPT-4's capabilities in thyroid ultrasound diagnosis and treatment recommendations using 109 cases. GPT-4 excelled in report structuring, clarity, and professional terminology but showed limitations in diagnostic accuracy. The chain of thought method enhanced interpretability, and the AI-generated reports were largely indistinguishable from human-written ones in a Turing Test. Zhang et al.  \cite{zhang2024berttcr} presented BertTCR, an advanced \ac{DL} framework for predicting cancer-related immune status via T cell receptor (TCR) repertoire analysis. BertTCR leverages a pre-trained protein-\ac{BERT} model to extract high-dimensional features, incorporating \ac{CNN}, multiple instance learning, and ensemble techniques to enhance accuracy. Validated on datasets for thyroid and lung cancer, it achieves notable \ac{AUC} improvements over existing methods. The framework's flexibility supports universal cancer detection and immune status assessment. BertTCR's findings emphasize its potential for early cancer detection, personalized medicine, and broad applications in immune-related diagnostics. Table \ref{table:7} presents a summary of the performance of various LLM-based \ac{TCD} methods across different metrics.

\color{black}

\section{Limitations and challenges} 
\label{sec5}

Recognizing the obstacles in integrating \ac{AI} solutions into healthcare practices, including infrastructural, regulatory, and cultural challenges, is essential. Highlighting the critical role of cross-disciplinary cooperation in seamlessly integrating \ac{AI} into healthcare systems, thereby maximizing its beneficial impacts on patient health outcomes.

Although \ac{AI} methodologies have shown promise in \ac{TC} diagnosis, they face challenges that hinder the development of efficient solutions, lead to increased expenses, and limit their broad application. For precise detection of \ac{TC}, it's essential to collect and securely consolidate all relevant data in a single repository, unless adopting \ac{FL} approaches, as Himeur et al. suggest \cite{himeur2023federated}. Following this, algorithms capable of identifying all forms of \ac{TC} must be developed. Comprehensive \ac{TCD} should include an extensive array of training and testing images, diagrams of nodules, and detailed classifications of nodule characteristics across different sizes, as Shah et al. recommend \cite{shah2023deep}. It's crucial for these datasets to be continually updated with data from \ac{MRI}, \ac{CT} scans, X-rays, and other clinical images to assess \ac{TC} accurately. The inclusion of demographic details such as race, ethnicity, gender, and age is also necessary. Establishing a centralized database accessible to all healthcare facilities for testing, validating, and implementing \ac{AI} algorithms on the collected data is critical, following Salazar et al.'s guidance \cite{salazar2019thyroid}. Additionally, a succinct overview of further limitations and challenges yet to be addressed is provided.

\noindent \textbf{{(a) Clean and sufficient labelled data to ensure accuracy:}} In \ac{TC} diagnosis, a major challenge is the lack of detailed, well-annotated datasets that thoroughly document the disease's incidence and progression. Elmore et al. \cite{elmore2021blueprint} highlight the difficulty in collecting and validating TC-related data due to the absence of comprehensive clinical databases. AI algorithms struggle with accurate \ac{TC} diagnosis because of limited labelled cases correlating with clinical outcomes, as noted by Park et al. \cite{park2021key}. Although large datasets are crucial for neural networks to produce accurate results, selective data incorporation during training is necessary to avoid harmful noise. Imaging modalities like \ac{CT} and \ac{MRI}, though available, are costly and not always accessible, as Ha et al. \cite{ha2021applications} point out. \ac{US} imaging, combined with physical exams, fine-needle aspiration biopsies, or radioisotope scans, is preferred for its cost-effectiveness and accessibility. However, Zhu et al. \cite{zhu2021generic} note that \ac{US} accuracy in distinguishing malignant from benign nodules can vary and images may be noisy. Cancerous cells in thyroid tissue are often a small fraction of the total dataset, leading to a skewed distribution that can impair AI detection performance, as observed by Yao et al. \cite{yao2022deepthy}. Researchers face challenges in developing algorithms to handle limited, noisy, sparsely annotated, incomplete, or high-dimensional samples efficiently. Annotation is crucial but time-consuming and costly, impacting AI algorithm precision due to inconsistent labelling, as discussed by Sayed et al. \cite{sayed2023time} and Yao et al. \cite{yao2022deepthy}.

\color{black}

\noindent\textbf{(b) Hyperparameters of DL models:}  Designing the effective \ac{DL} algorithm is crucial for overcoming different challenges, especially in diagnosing \ac{TC}. The task of precisely differentiating between malignant and benign tumours, is complex due to their significant similarities, as highlighted in the study by Wang et al. \cite{wang2022soft}. Addressing this challenge may require significantly increasing the number of DL layers for feature extraction. However, such an increase can lead to longer processing times, particularly with large datasets, which may delay timely cancer diagnoses for patients, as pointed out in the research by Lin et al. \cite{lin2021deep}.

\noindent\textbf{(c) Computation cost and storage limitations:} Pose notable hurdles in algorithm development. Time complexity, a key measure in algorithm evaluation, assesses computational complexity by approximating the count of basic operations performed and its relationship with the size of input data. Typically represented as $O(n)$, where $n$ is the size of the input, often measured by the bits required for its representation, this concept is thoroughly examined in the study by Al et al. \cite{al2021cost}. Particularly in \ac{AI} research related to {\ac{TCD}}, researchers are tasked with finding algorithms that offer a harmonious blend of accuracy and computational efficiency. Their goal is to develop algorithms that can quickly process large datasets while maintaining precise results. Furthermore, the extensive amount of data used in these algorithms sometimes exceeds the storage capabilities, an issue underscored in the research by Lin et al. \cite{lin2021deep}.

\noindent\textbf{(d) Data loss and Error vulnerabilities:}  The shift towards digital medical records is crucial, notably in cancer diagnosis using slide images. This latter facilitates the use of \ac{AI} for pathologic examinations \cite{dov2019thyroid}. Nonetheless, medical digitalization encounters specific challenges. There exists the danger of losing critical information during the digital conversion process and potential inaccuracies due to data compression methods applied in autoencoder algorithms. Thus, selecting the right digitalization technology is essential to ensure the preservation of data fidelity and authenticity \cite{halicek2019head}.  The subtle contrast between the thyroid gland and surrounding tissues complicates accurate analysis and diagnosis of \ac{TC}.
\color{black} Despite \ac{AI}'s inherent autonomy, it is prone to making errors. For example, training an algorithm with \acp{TCD} for identifying cancerous regions can lead to biased predictions if the training datasets are biased. These biases may then lead to a series of erroneous results, which could go unnoticed for a significant duration. Identifying and correcting the source of these errors, once recognized, can be a laborious process, as explored in the research by Karsa et al. \cite{karsa2020optimized}.

\noindent\textbf{(e) Unexplainable \ac{AI}:} The application of \ac{AI} in healthcare sometimes results in "black box" outcomes, where the decision-making process lacks transparency and sufficient justification. This lack of clarity can make healthcare practitioners question the dependability of the results, possibly leading to incorrect choices and treatments for patients with \ac{TC}. In essence, \ac{AI} systems can operate as black boxes, providing outcomes without explicit and comprehensible rationales, a concern highlighted in the research by Sardianos et al. \cite{sardianos2021emergence}.

\noindent\textbf{(f) Lack of cancer detection platform:} A significant barrier in detecting various cancers, including \ac{TC}, is the absence of platforms that facilitate the replication and evaluation of previous research. This gap presents a considerable challenge that hampers the assessment of \ac{AI} algorithms' performance, thereby hindering improvements \cite{abdolali2020automated}. The presence of online platforms that offer cutting-edge algorithms, extensive datasets, and expert insights is crucial for assisting healthcare practitioners, specialists, researchers, and developers in making the right decisions with reduced chances of error. Moreover, this kind of platforms are vital in augmenting clinical diagnoses, as they enable more thorough Examination and assessment \cite{masuda2021machine}.

\section{Future research directions} 
\label{sec6}
This segment delves into the anticipated developments of \ac{AI} in identifying \ac{TC}, scrutinizing forthcoming trends and advancements alongside their ethical repercussions. Ethical concerns encompass more than the immediate area of focus; issues regarding data privacy, responsibility, and fairness are also discussed. This part underscores research avenues poised to significantly improve \ac{TCD} classification and prediction.

\begin{figure}[t!]
\centering
\includegraphics[width=0.5\columnwidth]{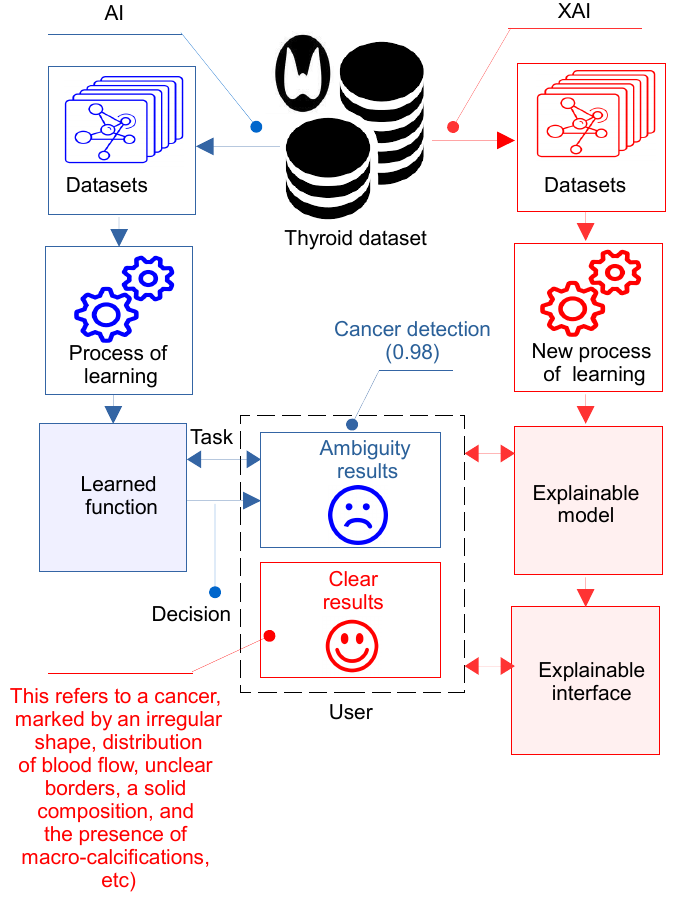}
\caption{Diagram of \ac{XAI} blocks {for \ac{TCD}}. }
\label{fig12}
\end{figure}

\noindent \textbf{(a) Employing XAI: } 
Integrating \ac{AI} into decision-making is crucial but faces challenges due to its complexity and lack of clarity. To mitigate these issues, \ac{XAI} seeks to make \ac{AI} models more transparent, more accurate and confident decisions. This is particularly vital in healthcare, where understanding the rationale behind \ac{AI}-generated outcomes is paramount. \ac{XAI} has been applied to \ac{TCD}, as evidenced in studies by Lamy et al. \cite{lamy2020intelligence}, and Poceviciute et al. \cite{poceviciute2020survey}. The differentiation between standard \ac{AI} and \ac{XAI} is showcased in Figure \ref{fig12}. Wildman et al.  \cite{wildman2019using} proposed an \ac{XAI} approach for detecting \ac{TC}, enhancing the confidence of healthcare practitioners in AI predictions. \ac{XAI} models clarify their reasoning, addressing the limitations associated with opaque "black box" algorithms. 

\noindent \textbf{(b) Using cloud, fog, and edge  computing: } 
The concept of edge networks merges edge computing with \ac{AI}, allowing \ac{AI} algorithms to operate closer to where data originates, a discussion brought forward by Sayed et al.  \cite{sayed2023edge}. This method enhances efficiency and cost-effectiveness for data-intensive applications, minimizing the requirement for extensive communication between patients and healthcare providers. By positioning data and storage closer to users in the healthcare field, this approach enables direct and swift access, a point highlighted by Alsalemi et al.  \cite{alsalemi2022innovative}. To further improve the detection of \ac{TC} within edge networks, fog computing is integrated. Fog computing introduces a distributed framework that bridges cloud computing and data generation sources, offering a versatile distribution of computational and storage capacities at key locations to boost overall system performance \cite{sayed2021intelligent}. Cloud computing acts as a pivotal facilitator for the efficient functioning of \ac{AI}-driven \ac{TCD} systems, offering readily available access to data storage, servers, databases, networks, and applications for healthcare professionals, contingent upon internet connectivity. This integrated approach has proven its worth in medical scenarios, such as in the \ac{TC} detection, as corroborated by several researches \cite{sufian2020survey,rajan2020fog}.

\noindent \textbf{(c) \Acf{DRL}: } \Ac{RL}, a branch of \ac{ML} \cite{kheddar2024reinforcement}, enables agents to navigate and make decisions in evolving environments by engaging in a learning cycle of trial and error, observation, and interaction. The interest in leveraging \ac{RL} for diagnosing untreatable diseases and enhancing the support for medical decision-making processes has grown recently. For example, Balaprakash et al.  \cite{balaprakash2019scalable} apply \ac{RL} in cancer data classification, whereas Li et al.  \cite{li2020deep} explore the use of deep \ac{RL} for lymph node dataset segmentation. In this approach, pseudo-ground truths are created using RECIST-slices, facilitating the simultaneous tuning of lymph node bounding boxes through the collaborative efforts of a segmentation network and a policy network.

\noindent \textbf{(d) \Acf{DTL}: }
\Ac{DTL} is recognized as an effective approach to reduce overfitting and improve the precision of diagnostic tools \cite{kerdjidj2023uncovering,himeur2023video}. This technique applies the knowledge acquired from one domain to solve related issues in another, such as shortening the duration of training and minimizing the amount of data needed \cite{kheddar2023deep,sayed2023time}. It is particularly useful in diagnosing \ac{TG}. For example, the Enhance-Net model, described in \cite{narayan2023enhance}, could act as a foundational model to boost the efficacy of a targeted \ac{DL} model aimed at analyzing medical images in real-time. Furthermore, in \cite{liu2017classification}, the research focuses on identifying pertinent characteristics of benign and malignant nodules using \acp{CNN}. By transferring insights from generic data to a dataset of \ac{US} images, they achieve a fusion of hybrid semantic deep features. The application of transfer learning has also proven beneficial in categorizing images of \acp{TN}, as shown in \cite{song2019ultrasound}.

\noindent \textbf{(i) \Acf{FL}:} \acs{FL} has gained traction in healthcare applications due to its capacity to enhance patient data privacy across different healthcare settings \cite{himeur2023federated}. The influence of environmental factors on human health, which can subsequently impact economic stability, is substantial. An increase in the incidence of thyroid gland disorders has been observed across diverse populations. \ac{ML} plays a crucial role in addressing such health issues by leveraging collected data to train models capable of foreseeing severe health conditions. Considering the critical need for maintaining the confidentiality of patient information among various health institutions, \ac{FL} stands out as an optimal framework for these purposes. Lee et al. \cite{lee2021federated} conducted a study comparing the effectiveness of FL with five traditional DL techniques in analyzing and detecting {\ac{TC}}.

\noindent \textbf{(e) \Acf{PS}: } Accurately identifying and segmenting objects with varied and intersecting features continues to be a significant hurdle, especially in the medical field. To tackle this issue, several scholars have developed holistic and unified segmentation methods \cite{elharrouss2021panoptic}. Panoptic segmentation has received considerable attention, merging the principles of instance and semantic segmentation to detect and delineate objects efficiently. Semantic segmentation involves the classification of each pixel into distinct categories, whereas instance segmentation focuses on delineating individual object instances. \ac{AI} has been applied to this framework through either supervised or unsupervised instance segmentation learning techniques, making it highly applicable to medical scenarios \cite{yu2020deep}.

\noindent \textbf{(f) IoMIT and 3D-TCD:}  The \ac{IoMIT} has gained substantial interest in the healthcare industry in recent years. \ac{IoMIT} seeks to advance the quality of healthcare services and minimize treatment expenses by facilitating the exchange of medical information between patients and healthcare providers via interconnected devices equipped with wireless communication technology. An instance of such integration is showcased in \cite{ivanova2018artificial}, where an \ac{AI}-enhanced solution for the preemptive identification of \ac{TC} within the \ac{IoMIT} paradigm is introduced. This method employs \ac{CNN} to refine the distinction between benign and malignant nodules, aiming ultimately at life preservation. Additional investigations pertinent to \ac{IoMIT} \cite{borovska2018internet}. 2D \ac{US} is a prevalent technique for evaluating \acp{TN}, yet its static imagery might not fully capture the nodules' complex structures. Consequently, there's a growing interest in utilizing three-dimensional (3D) \ac{US}, which offers a holistic view of the lesion by reconstructing nodule characteristics, thereby facilitating enhanced discrimination between different diagnostic categories \cite{seifert2023optimization}. The capability of 3D \ac{US} to analyze intricate growth patterns, edges, and forms from various perspectives and depths allows for a more accurate assessment of \acp{TN}' morphological features compared to 2D \ac{US}. 

\noindent \textbf{(g) \ac{AI}-based thyroid surgical techniques: } As surgical practices face complex challenges, the essential role of \ac{AI}-driven robotic assistance is becoming increasingly recognized. \ac{AI} has the capability to navigate clinical intricacies by processing and leveraging large volumes of data, offering decision-making support with a precision that rivals that of medical experts \cite{pakkasjarvi2023artificial}. Businesses are \ac{AI} into surgical operations through the development of \ac{AI} systems and the deployment of robots to aid surgeons in the operating room. These robots fulfil various functions, such as managing surgical tools, handling potentially contaminated materials and medical waste, conducting remote patient monitoring, and compiling patient information including electronic health records, vital signs, lab results, and video documentation \cite{bodenstedt2020artificial}. It is therefore vital for surgeons to develop a comprehensive understanding of \ac{AI} and its potential impacts on healthcare. While \ac{AI}-enabled robotic surgery is still emerging, fostering interdisciplinary collaboration can accelerate the progress of \ac{AI} technology, thereby improving surgical outcomes \cite{lee2020evaluation}.

\noindent \textbf{(h) Recommender systems (RSs): } The vast amount of information produced by online medical platforms and electronic health records presents a challenge for \ac{TC} patients seeking specific and accurate data \cite{himeur2021survey}. Additionally, the substantial costs associated with healthcare data management can complicate the task of physicians handling a broad spectrum of patients and treatment alternatives. The implementation of RS has been suggested as a solution to improve decision-making within healthcare, reducing the load on both patients and oncologists \cite{areeb2023filter, varlamis2022smart}. Incorporating RS into digital health facilitates tailored recommendations, precise evaluation of large data sets, and stronger privacy measures, leveraging the capabilities of \ac{AI} and ML technologies \cite{atalla2023intelligent}.

\noindent \textbf{(k) Image and video compression, and denoising for {\ac{TCD}}: } The use of medical image and video compression plays a pivotal role in enhancing the detection and diagnosis of cancer, leveraging the advancements in digital imaging and telecommunications. This technological advancement allows for the efficient storage and transmission of high-resolution diagnostic images such as X-rays, MRIs, and CT scans, which are critical in identifying malignant tumors at early stages. Compression algorithms, both lossless and lossy, are meticulously designed to ensure that the integrity of the diagnostic information is maintained, making it possible for radiologists and oncologists to discern fine details crucial for accurate diagnosis. Furthermore, the reduced file sizes facilitate quicker transfer speeds across networks, enabling real-time collaboration and consultation among healthcare professionals worldwide, thereby significantly improving the speed and accuracy of cancer diagnosis. This is especially vital in remote or resource-limited settings where access to high-quality healthcare and specialist consultations might be restricted, thus democratizing the access to crucial diagnostic services and improving patient outcomes \cite{habchi2023new, beladgham2019medical,habchi2022improving,habchi2021ultra, habchi2016rgb}. In addition, applying denoising techniques to medical images before training can substantially improve the accuracy of TCD classification and decision-making processes \cite{boucherit2025reinforced}.

\noindent \textbf{(l) Features selection: } 
The \ac{IG} method is useful in simplifying the classification of medical images. Researchers are encouraged to explore its usefulness for detecting \ac{TC} by identifying the most informative features that distinguish between malignant and benign \acp{TN}. The process begins with data collection, where a comprehensive dataset containing relevant features, such as patient demographics, ultrasound characteristics, biopsy results, genetic markers, and blood test results, is gathered. Each instance in the dataset is labelled as benign or malignant based on definitive diagnostic methods. In the preprocessing stage, the data is cleaned by handling missing values and outliers and normalizing if necessary. Feature engineering may also be performed to create new features that enhance predictive power. The \ac{IG} for each feature is then calculated, measuring how much it contributes to the classification. Features with high \ac{IG} are considered more informative and are used to build a predictive model for \ac{TC} detection. This method helps in selecting the most relevant features, thereby improving the accuracy and efficiency of the diagnostic process.

\noindent \textbf{(m) Generating synthetic datasets:}  To advance \ac{TCD}, future research should focus on enhancing dataset quality and diversity. Developing well-annotated datasets remains a challenge, which can be addressed through innovative techniques such as synthetic data generation, data augmentation, and multi-modal integration. Future work could explore more effective synthetic data generation methods, including enhanced \acp{GAN} and variational \acp{DAE}, to create diverse, high-quality datasets that improve diagnostic accuracy and represent rare cancer subtypes \cite{gangwal2024current}. Further investigation into advanced data augmentation strategies, such as complex image transformations and domain adaptation, could enhance model generalization by expanding dataset variability \cite{kumar2024image}. Additionally, multi-modal integration, combining imaging, genomics, and clinical data, holds promise for improving robustness and predictive performance through deep learning models and novel fusion strategies.

\noindent \textbf{(n) Employing SSL:} \ac{SSL} offers a promising avenue for feature extraction in \acp{TCD}, particularly when working with large volumes of unannotated medical images. By leveraging unlabeled data, \ac{SSL} techniques can learn meaningful and discriminative representations, reducing the dependency on extensive manual annotation efforts. This approach has shown the potential to improve the robustness and generalization of diagnostic models by capturing complex patterns within medical imaging data. Incorporating \ac{SSL} strategies could further enhance the development of automated diagnostic tools, especially in data-limited scenarios. Therefore, a dedicated discussion of \ac{SSL} methods, including their applications and potential benefits in the \ac{TCD} field, has been added to provide a comprehensive overview of emerging advancements. Recent studies have highlighted the effectiveness of \ac{SSL} in medical image classification. For instance, \cite{huang2023self} discusses various \ac{SSL} strategies and their applications in medical imaging, emphasizing their potential to improve diagnostic performance. Additionally, research published in \cite{wolf2023self} explores \ac{SSL} pre-training approaches, such as contrastive and masked modelling, demonstrating their superiority over traditional supervised methods in medical imaging tasks.

\color{black}

\section{Conclusion}  
\label{sec7}
This investigation delves deeply into \ac{ML} and \ac{DL}, highlighting their growing prominence due to their enhanced precision over other methods. It comprehensively reviews various algorithms and training models, discussing their benefits and drawbacks. Specifically, \ac{DL} techniques are celebrated for their application in a myriad of real-world scenarios, notably for their generalization capabilities and resilience to noise. Nevertheless, significant challenges obstruct the full adoption of \ac{DL} in detecting \ac{TC}, with the lack of clean data and appropriate platforms being primary concerns. Tackling these data challenges with detailed precision is essential for creating effective and robust models for detecting more complex cancer stages.

Future research should aim at overcoming these hurdles and improving \ac{TCD} classification and prediction methods. This study highlights the urgent need for increased research focus on \ac{TC} diagnostics to match the high precision expectations of healthcare practitioners. While cancer detection in two or three dimensions is progressing, the limited expertise in handling various geometric transformations and multi-dimensional data compromises the accuracy of diagnosing life-threatening diseases. Therefore, it is vital to innovate in distinguishing between cancerous nodule sizes. Such innovations could significantly speed up treatment, improve diagnostic precision, foster proactive epidemiological tracking, and reduce death rates. Novel technologies like  \ac{XAI}, edge computing, \ac{DTL}, \ac{RL}, \ac{FL} for privacy-preserving mechanisms, and remote sensing are paving new paths in AI-based \ac{TCD} research. These developments are crucial for medical professionals, simplifying the diagnostic process, reducing detection times, and enhancing patient confidentiality. Future research will explore the impact of these advanced technologies further. The objective is to create a major transformation in cancer detection approaches by crafting advanced, privacy-focused technologies for the identification of \ac{TC} and extending into domains like Telehealth.

\begin{table}[ht!]
{\small \section*{Acronyms and Abbreviations}}
\leftskip=-4cm 
\begin{multicols}{3}
\scriptsize
\begin{acronym}[AWGNFFG]  
\acro{AC}{active contour}
\acro{AI}{artificial intelligence}
\acro{AMFFM}{adaptive multi-scale feature fusion module}
\acro{ANN}{artificial neural network} 
\acro{ATC}{anaplastic thyroid carcinoma}
\acro{ATM}{assembled Transformer  module}
\acro{AUC}{area under curve}
\acro{BA}{bootstrap aggregation}
\acro{BERT}{bidirectional encoder representations from Transformer}
\acro{Bi-LSTM}{bi-directional LSTM}
\acro{BPAT-UNet}{boundary-preserving assembly tansformer UNet}
\acro{BPSM}{boundary point supervision module}
\acro{CAD}{computer-aided diagnosis}
\acro{CEUS}{contrast-enhanced ultrasound}
\acro{CFS}{correlation-based feature selection}
\acro{CNN}{convolutional neural network}
\acro{Corr}{correlation}
\acro{CT}{computed tomography} 
\acro{DAE}{denoising autoencoder}
\acro{DCNN}{deep convolutional neural network}
\acro{DDTI}{digital database thyroind image}
\acro{DL}{deep learning}
\acro{DNN}{deep neural network}  
\acro{DRL}{deep reinforcement learning}
\acro{DT}{decision tree}
\acro{DTCW}{double-tree complex wavelet transform}
\acro{DTL}{deep transfer learning}
\acro{DWT}{discrete wavelet transform}
\acro{ELM}{extreme learning machine} 
\acro{EM}{ensemble method}
\acro{ER}{error rate}
\acro{FB}{feature bagging}
\acro{FCM}{fuzzy c-means}
\acro{FL}{federated learning}
\acro{FNAB}{fine-needle aspiration biopsy}
\acro{FTC}{follicular thyroid carcinoma}
\acro{GAN}{generative adversarial network}
\acro{GEO}{gene expression omnibus}
\acro{GLCM}{gray-level co-occurrence matrix}
\acro{GPT}{generative pre-trained Transformer }
\acro{GPU}{graphics processing unit}
\acro{HOG}{histogram of oriented gradient}
\acro{ICA}{independent component analysis}
\acro{IG}{information gain}
\acro{IoMIT}{internet of medical imaging thing}
\acro{JSI}{jaccard similarity index}
\acro{KM}{K-means}
\acro{KNN}{k-nearest neighbors}
\acro{LBP}{local binary patterns}
\acro{LLM}{large language model}
\acro{LMSE}{laplacian mean square error}
\acro{LR}{logistic regression}
\acro{LSTM}{long-short-term-memory}
\acro{MAE}{mean absolute error}
\acro{ML}{machine learning}
\acro{MLP}{multilayer perceptron}
\acro{MRI}{magnetic resonance imaging}
\acro{MRM}{microRNA regulatory module}
\acro{MRR}{mean reciprocal rank}
\acro{MSE}{mean square error}
\acro{MTC}{medullary thyroid carcinoma}
\acro{NAE}{normalized absolute error}
\acro{NCC}{normalized cross-correlation}
\acro{NCDR}{national cancer data repository}
\acro{NLP}{natural language processing}
\acro{NVF}{Noise visibility function}
\acro{OCT}{optical coherence tomography}
\acro{PCA}{principal component analysis}
\acro{PFCM}{possibilistic fuzzy c-means}
\acro{PLCO}{prostate, lung, colorectal, and ovarian} 
\acro{PM}{probabilistic models}
\acro{PS}{panoptic segmentation}
\acro{PSNR}{peak signal to noise ratio} 
\acro{PTC}{papillary carcinoma}
\acro{RA}{relevance analysis}
\acro{RBM}{restricted Boltzmann machine}
\acro{RF}{random forest}
\acro{RL}{reinforcement learning}
\acro{RMSE}{root mean square error} 
\acro{RNN}{recurrent neural network}
\acro{ROC}{receiver operating characteristic}
\acro{ROI}{region of interest}
\acro{SC}{structural content}
\acro{SD}{standard deviation}
\acro{SL}{supervised learning} 
\acro{SSL}{self-supervised learning}
\acro{SVM}{support vector machine}
\acro{TC}{thyroid cancer}
\acro{TCD}{thyroid cancer diagnosis}
\acro{TCGA}{cancer genome atlas}
\acro{TCL}{traditional classification}
\acro{TD}{thyroid disease}
\acro{TDD}{thyroid disease diagnosis} 
\acro{TFD}{discrete Fourier transforms}
\acro{TG}{thyroid gland} 
\acro{TI}{thyroid inputs}
\acro{TIRADS}{thyroid imaging reporting and data system}
\acro{TL}{transfer learning}
\acro{TN}{thyroid nodule}
\acro{TT}{thyroid targets} 
\acro{UCI}{university of California, Irvine}
\acro{US}{ultrasound}
\acro{USL}{unsupervised learning}
\acro{ViT}{vision Transformer }
\acro{VOE}{volumetric overlap error}
\acro{VSNR}{visual signal to noise ratio}
\acro{WSI}{whole slide histopathological images}
\acro{WSNR}{weighted signal-to-noise ratio}
\acro{XAI}{explainable artificial intelligence}
\acro{XGBoost}{gradient tree boosting}
\end{acronym}

\end{multicols}
\end{table}

\section*{Data availability}
Data will be made available on request.

\section*{Conflict of Interest}
The authors declare no conflicts of interest.

\section*{References}
\leftskip=-4cm 
\bibliography{references}

\end{document}